%% file: main.tex
\documentclass{article}

\usepackage{PRIMEarxiv}

\usepackage[utf8]{inputenc} % allow utf-8 input
\usepackage[T1]{fontenc}    % use 8-bit T1 fonts
\usepackage{hyperref}       % hyperlinks
\hypersetup{hidelinks}
\usepackage{url}            % simple URL typesetting
\usepackage{booktabs}       % professional-quality tables
\usepackage{amsfonts}       % blackboard math symbols
\usepackage{nicefrac}       % compact symbols for 1/2, etc.
\usepackage{microtype}      % microtypography
\usepackage{graphicx}       % graphics
\usepackage{array}

\input{math_commands.tex}

\title{Diff-ID: Identity Consistent Facial Image Generation\\
and Morphing via Diffusion Models}

\author{
  Taimoor Rizwan, Sara Atito, Muhammad Awais, Josef Kittler \\
  Centre for Vision, Speech and Signal Processing (CVSSP) \\
  University of Surrey \\
  Guildford, United Kingdom \\
  \texttt{\{tr00564, sara.atito, muhammad.awais, j.kittler\}@surrey.ac.uk} \\
  \AND
  Zhenhua Feng \\
  School of Artificial Intelligence and Computer Science \\
  Jiangnan University \\
  Wuxi, China \\
  \texttt{fengzhenhua@jiangnan.edu.cn} \\
}

\begin{document}
\maketitle

\begin{abstract}
Generative diffusion models have revolutionized facial image synthesis, yet robust identity preservation in high resolution outputs remains a critical challenge. This issue is especially vital for security systems, biometric authentication, and privacy sensitive applications, where any drift in identity integrity can undermine trust and functionality. We introduce Diff-ID, a diffusion based framework that enforces identity consistency while delivering photorealistic quality. Central to our approach is a custom 210K image dataset synthesized from CelebA-HQ, FFHQ, and LAION-Face and captioned via a fine tuned BLIP model to bolster identity awareness during training. Diff-ID integrates ArcFace and CLIP embeddings through a dual cross attention adapter within a fine tuned Stable Diffusion UNet. To further reinforce identity fidelity, we propose a pseudo discriminator loss based on ArcFace cosine similarity with exponential timestep weighting. Experiments on held out and unseen faces show that Diff-ID does not exceed InstantID in raw ArcFace Face Similarity, but achieves substantially lower FID and the strongest FIQ based identity--realism trade off among the evaluated methods. We also present a unified DDIM based morphing pipeline that enables qualitative facial interpolation without per identity fine tuning. We further argue that identity preservation and photorealism should be evaluated jointly rather than in isolation, as high identity similarity alone does not guarantee realistic outputs. To make this trade off explicit, we report Face Image Quality (FIQ) as a complementary ratio based score that combines identity similarity and perceptual realism while keeping FS and FID as the primary metrics.

\end{abstract}


\section{Introduction}
Generative diffusion models have recently emerged as a leading paradigm for high fidelity image synthesis, outperforming earlier techniques such as Generative Adversarial Networks (GANs) and Variational Autoencoders (VAEs) in training stability, diversity, and output quality~\cite{ho2020denoising, rombach2022high}. By iteratively refining noisy latent representations through a learned denoising process, diffusion based frameworks can generate hyper realistic images that capture intricate textures, complex structures, and subtle semantic variations.

The success of diffusion models spans diverse applications: from artistic content creation and virtual reality to medical imaging and scientific visualization, where photorealism and anatomical precision are critical~\cite{nichol2021glide, saharia2022photorealistic}. In the domain of facial image synthesis, diffusion methods have demonstrated impressive visual quality, yet they exhibit limitations when precise control over identity attributes is required. In particular, subtle identity drift, manifesting as changes in facial geometry, expression, or distinguishing features, can undermine the trustworthiness of generated outputs in scenarios demanding strict identity fidelity.

Maintaining robust identity consistency is a multifaceted challenge: models must preserve core identity defining features (for example bone structure, eyes, and facial contours) under varying manipulations (for example expression, pose, lighting). However, many current diffusion based personalization methods~\cite{liu2023ip,zhang2023photomaker,chen2023instantid,zhang2023controlnet} emphasize global attribute changes, such as clothing styles or accessories, and may trade off identity fidelity, photorealism, and fine grained semantic alignment in different ways.

This identity control gap poses a significant barrier for security sensitive applications, including biometric authentication, forensic analysis, and privacy preserving data generation, where any deviation from the true identity can have severe consequences. Synthetic morphing pipelines, used for adversarial robustness testing and privacy aware dataset augmentation, further demand seamless identity interpolation without per identity retraining or checkpoint swapping. 

Beyond methodological motivation, identity consistent and photorealistic face generation is essential for several practical application scenarios. A primary example arises from data protection and privacy regulations such as GDPR, which increasingly restrict the use of real biometric data for training and evaluation. In this context, synthetic but photorealistic identities provide a viable alternative, enabling the development and benchmarking of face recognition systems without exposing real individuals. Crucially, such synthetic identities must preserve realistic facial structure while exhibiting diversity in pose, illumination, resolution, expression, and hairstyle in order to reflect real operational conditions.

A second important scenario is dataset augmentation. Modern face recognition systems require large and diverse training corpora, yet collecting balanced samples across capture conditions and demographics remains challenging. Identity preserving generative models enable controlled augmentation by synthesizing realistic variations of a subject without compromising identity coherence. Finally, identity consistent morphing provides a stepping stone toward morph attack generation and detection, where realistic blends of identities are required to study vulnerabilities of biometric systems. In all these scenarios, the simultaneous preservation of identity and photorealism is not optional but essential, motivating the design of Diff-ID.

To address these challenges, we propose \textbf{Diff-ID}, a unified diffusion framework explicitly designed for identity aware facial image synthesis and morphing. Our contributions are fivefold:
\begin{enumerate}
\item \textbf{Custom Identity Centric Dataset}: We curate a 210K image corpus by blending and captioning CelebA-HQ, FFHQ, and LAION-Face images with a fine tuned BLIP model, ensuring comprehensive coverage of facial variations and semantic contexts.
\item \textbf{Dual Cross Attention Adapter}: We develop a lightweight adapter that fuses identity embeddings from ArcFace with semantic embeddings from CLIP within a fine tuned Stable Diffusion UNet, supporting competitive identity preservation while improving photorealistic visual quality in our evaluation. Our framework does not aim to provide explicit attribute editing or fine grained attribute control.
\item \textbf{Pseudo Discriminator Identity Loss}: We introduce an ArcFace based identity loss with exponential dynamic weighting over diffusion timesteps, reinforcing identity coherence throughout the denoising process.
\item \textbf{Unified Morphing Pipeline}: Leveraging DDIM inversion and joint embedding plus latent interpolation, Diff-ID produces smooth qualitative morphing trajectories between arbitrary face pairs without additional model fine tuning or multiple checkpoints.
\item \textbf{Complementary Identity--Realism Evaluation Score}: We report the Face Image Quality (FIQ) score as a secondary ratio based indicator that couples ArcFace similarity with Fréchet Inception Distance, helping summarize the identity--realism trade off while preserving FS and FID as separate primary measurements.

\end{enumerate}

We validate Diff-ID on both held out and unseen high resolution facial datasets. The results show competitive ArcFace similarity, the lowest Fréchet Inception Distance (FID), and the highest FIQ among the evaluated baselines; importantly, Table~\ref{tab:evaluation} shows that InstantID retains the highest raw Face Similarity. These findings support Diff-ID as a strong FIQ based identity--realism trade off model rather than as the best method under raw identity similarity alone.

% ----------------------------------------------------------------------
% 2. Related Work
% ----------------------------------------------------------------------

\section{Related Work}
\label{sec:relatedworks}
Diff-ID builds on two converging lines of research: general diffusion based image synthesis and specialized identity preserving adaptations. We first review foundational text to image diffusion frameworks, then examine approaches for embedding based identity control, attention mechanisms, adapter modules, and morphing techniques.

\subsection{Text to Image Diffusion Models}
Text to image diffusion models have rapidly become the dominant paradigm in generative image synthesis by combining iterative denoising processes with powerful text encoders. Early transformer based approaches, such as DALL\,$\cdot$E~\cite{ramesh2021zero}, CogView~\cite{ding2021cogview}, and Make A Scene~\cite{gal2022make}, demonstrated rich text image correlations via discrete tokenization, but they incurred substantial compute and latency costs at ultra high resolutions due to quadratic self attention complexity~\cite{ramesh2021zero}.

Diffusion models overcome these bottlenecks by operating in continuous latent spaces and gradually transforming noise into coherent images. Song and Ermon formalized this process via stochastic differential equations, showing that score matching on Gaussian perturbations recovers complex data distributions~\cite{song2021score}. Ho et al.'s denoising diffusion probabilistic models then introduced practical noise schedules for stable, end to end training~\cite{ho2020denoising}.

Subsequent advances improved both fidelity and efficiency. GLIDE proposed classifier free guidance, enabling a simple trade off between fidelity and diversity without an external classifier~\cite{nichol2021glide}. DALL\,$\cdot$E 2 adopted a hierarchical two stage diffusion, first generating CLIP embeddings from text, then super resolving back to image space, which further boosted sample quality~\cite{ramesh2022hierarchicaltextconditionalimagegeneration}. Imagen achieved state of the art FID on COCO 30k by scaling up text encoders and carefully tuning noise schedules~\cite{saharia2022photorealistic}.

Latent Diffusion Models marked another leap by performing denoising in a compressed latent space, reducing memory and computation while retaining perceptual detail~\cite{rombach2022high}. Stable Diffusion, an open source latent diffusion model at \(512\times512\) resolution, democratized these techniques and inspired numerous extensions for style, structure, and spatial conditioning~\cite{zhang2023adding}. Despite these successes in text driven generation, vanilla diffusion pipelines do not include mechanisms to anchor outputs to a specific face identity. As a result, attribute manipulations, such as altering expression or pose, can inadvertently shift identity features, leading to drift. This limitation underscores the need for identity aware diffusion frameworks like Diff-ID.

\subsection{Zero Shot Embedding Approaches}
Zero shot embedding techniques have recently emerged as an efficient way to steer pretrained text to image diffusion models toward identity preserving outputs without per subject fine tuning. Many methods build on large scale vision language encoders such as CLIP~\cite{radford2021learning} to extract semantic image representations, and then inject these into the generation process. IP Adapter~\cite{liu2023ip} uses a decoupled cross attention mechanism that maintains separate query, key, and value projections for text and image embeddings, allowing detailed visual prompts to be fused without updating the backbone model weights. Similarly, InstantID~\cite{chen2023instantid} leverages a frozen face recognition encoder to produce identity embeddings on the fly, which are concatenated with text tokens at each denoising step to guide the model toward the correct subject.

Despite their efficiency, CLIP based zero shot embeddings can lack the fine grained discriminability needed for strict identity retention. In particular, they may fail to capture subtle but crucial facial characteristics, such as jawline contour, eye shape and spacing, nose bridge structure, lip curvature, and overall facial bone structure, leading to identity drift under challenging poses or lighting conditions~\cite{chen2023instantid,liu2023ip}. To address this, Diff-ID integrates high precision ArcFace embeddings~\cite{deng2019arcface}, which are explicitly trained with an additive angular margin objective to maximize inter class separability and preserve intra class consistency, into a frozen Stable Diffusion pipeline. By concatenating these embeddings with the model intermediate feature maps and applying a lightweight adapter layer, Diff-ID is designed to retain identity specific attributes, such as precise jawline geometry, characteristic eye proportions, nose bridge angle, and unique lip shape, while improving visual realism across the evaluated generation tasks.

\subsection{Decoupled Cross Attention Mechanisms}
Decoupled cross attention is a technique that separates the processing of text and image inputs, improving how identity and attribute features are integrated during image generation. Models like IP Adapter~\cite{liu2023ip} and InstantID~\cite{chen2023instantid} utilize decoupled cross attention to independently attend to textual and visual inputs, aligning generated content more precisely with the text description. This mechanism is particularly beneficial for preserving semantic alignment in the outputs, allowing models to generate contextually relevant images based on text prompts.

However, while decoupled cross attention enhances attribute alignment, it alone does not ensure identity consistency, particularly in models reliant solely on CLIP embeddings. In cases where detailed identity features are critical, this limitation can lead to identity drift and compromise realism. Diff-ID explores an alternative design in which existing cross attention layers are tuned together with a lightweight identity adapter, yielding competitive identity similarity and improved realism in our experiments. While more advanced semantic or accessory control may still require additional modules such as ControlNet, our results suggest that a simpler adapter based design can provide a favorable identity--realism trade off without adding decoupled attention modules.

\subsection{Stacked ID Embeddings for Identity Fidelity}
A promising direction for identity preservation in text to image diffusion is the use of stacked identity embeddings, where multiple reference images of the same individual are jointly encoded to form a richer identity representation. PhotoMaker~\cite{zhang2023photomaker} exemplifies this strategy by first extracting per image identity features via a pretrained face encoder, then concatenating these into a single stacked embedding that is injected into each denoising step through a dual cross attention mechanism. By combining class based attributes (for example gender or age) with individual specific embeddings, PhotoMaker maintains consistent identity cues, such as jawline structure, eye spacing, and hairstyle, across variations in pose, expression, and lighting~\cite{zhang2023photomaker}.

This multi image fusion leads to substantial gains in identity fidelity. Similar ideas have appeared in related work on multi view face synthesis, where aggregating embeddings from different angles improves three dimensional consistency and realism~\cite{deng2019uv}. However, the stacked identity approach incurs significant computational cost: training an image encoder from scratch on large scale face datasets (for example MS Celeb 1M~\cite{guo2016ms}) and managing multiple high dimensional embeddings per subject can slow both training and inference.

In contrast, Diff-ID sidesteps these overheads by leveraging off the shelf ArcFace embeddings~\cite{deng2019arcface} within a frozen Stable Diffusion backbone~\cite{rombach2022high}. Rather than stacking many vectors, Diff-ID projects a single, high precision identity vector into the model intermediate feature space via a lightweight adapter. A dual cross attention block then fuses identity and textual attributes in parallel, preserving fine grained features, such as nose bridge angle and lip curvature, without the need for multi view encoding or encoder retraining. This design yields competitive identity similarity while reducing memory footprint and computational latency, and the experiments emphasize its advantage in the combined identity--realism trade off.

\subsection{Diffusion Based Morphing and Synthetic Identity Generation}
Interpolation and morphing between identities have long been explored with GANs via latent space operations. Early works such as StyleGAN~\cite{karras2019style} demonstrated smooth identity interpolation by spherical linear interpolation between learned style vectors, while subsequent methods such as StyleFlow~\cite{abdal2021styleflow} and GANSpace~\cite{harkonen2020ganspace} provided fine grained, attribute aware traversals in the GAN latent manifold. However, GAN based interpolation often suffers from mode collapse and fixed model checkpoints per identity or attribute setting.

Diffusion models offer an alternative morphing paradigm. DiffMorpher~\cite{lee2024diffmorpher} performs latent space interpolation by inverting two target images into DDIM embeddings and linearly blending these codes during reverse diffusion. To achieve sharp identity transitions, DiffMorpher relies on per identity low rank adaptation fine tuning~\cite{hu2021lora} and multiple diffusion checkpoints, which incurs substantial compute and engineering overhead.

In contrast, Diff-ID studies morphing within the same unified adapter used for identity conditioned generation. During DDIM inversion, we fuse ArcFace identity embeddings~\cite{deng2019arcface} with CLIP text embeddings~\cite{radford2021learning} through a dual cross attention block. This semantically conditioned noise injection produces qualitative latent trajectories for smooth transitions without external fine tuning or checkpoint swapping. As a result, Diff-ID provides a single model basis for continuous identity morphing, privacy preserving face blending, and future adversarial robustness testing, while the present paper evaluates this morphing component primarily through visual examples and endpoint similarities.

% ----------------------------------------------------------------------
% 3. Methodology
% ----------------------------------------------------------------------

\section{Methodology}
\label{sec:methodology}

\subsection{Data Preparation and Curation}
To support identity consistent and semantically rich facial generation, we curated a large scale dataset composed of images from three well known high resolution facial datasets: CelebA-HQ, FFHQ, and LAION-Face. All images were resized to a uniform resolution of 512 to ensure consistency during model training.
\begin{figure}[ht]
    \centering
    \includegraphics[width=0.5\columnwidth]{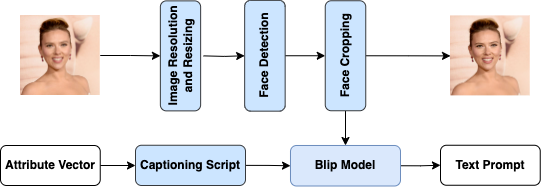}
    \caption{Data pipeline used to curate and caption the identity centric dataset.}
    \label{fig:Captioningdiagram}
\end{figure}

Unlike CelebA-HQ, which provides 40 annotated binary attributes per image, FFHQ and LAION-Face lacked descriptive captions. To address this, we developed a custom script that converts CelebA-HQ attribute vectors into structured sentence style captions. We then fine tuned a BLIP model~\cite{li2022blip} on this annotated subset to generate high quality captions for FFHQ and LAION-Face images.

To maintain data quality and semantic consistency, we removed images with occlusions (for example sunglasses, hats), low quality images, and any samples undetectable by the InsightFace library \cite{ren2023pbidr,guo2021sample,gecer2021ostec,an_2022_pfc_cvpr,an_2021_pfc_iccvw,deng2020subcenter,Deng2020CVPR,guo2018stacked,deng2018menpo,deng2018arcface}. This rigorous cleaning and captioning process resulted in a dataset of approximately 210{,}000 images, each paired with a descriptive and unique caption, which enables more expressive and semantically descriptive prompts while maintaining identity fidelity, without providing explicit attribute control.

\subsection{Overview of Diff-ID}
Diff-ID is a diffusion framework built on a fine tuned Stable Diffusion UNet \cite{rombach2022high} that is designed to encourage identity consistency in high resolution facial synthesis. At its core is a lightweight dual cross attention adapter that fuses high precision ArcFace identity embeddings~\cite{deng2019arcface} with CLIP semantic embeddings~\cite{radford2021learning}, anchoring generated outputs to the target identity. This design aims to preserve identity specific attributes, such as facial bone structure, eye spacing, nose shape, and lip contour, during denoising, while still allowing CLIP driven semantic prompts to provide broad contextual guidance within identity constraints.

To reinforce identity fidelity throughout the diffusion process, we introduce a pseudo discriminator identity loss: an ArcFace cosine similarity term with exponential timestep weighting that dynamically emphasizes the preservation of identity cues as noise levels decrease.
\begin{figure*}[ht]
    \centering
    \includegraphics[width=1.0\textwidth]{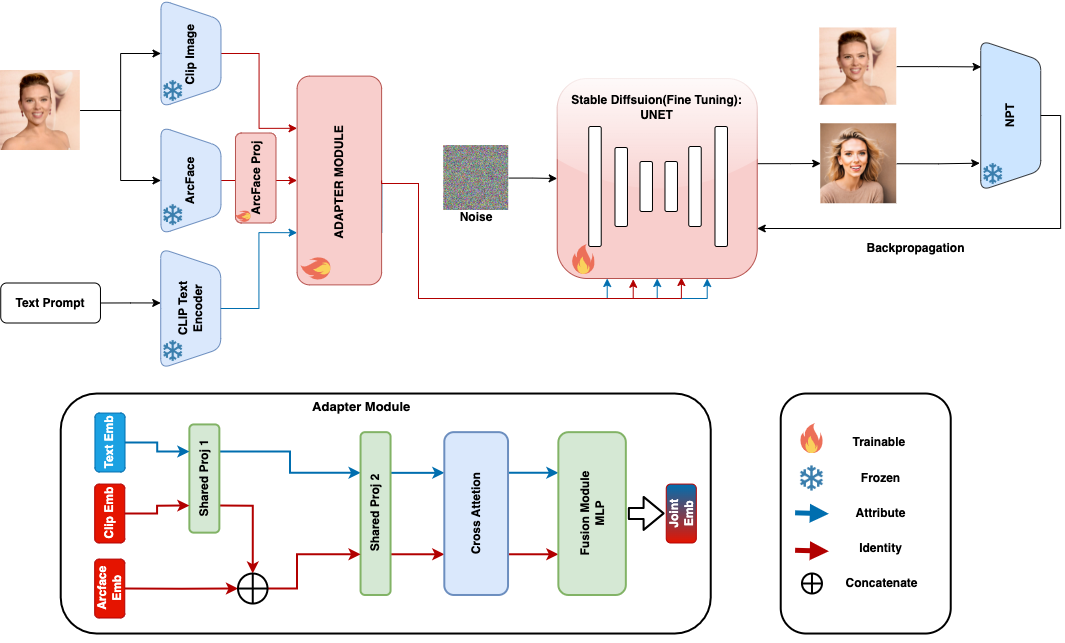}
    \caption{Diff-ID architecture. We extract semantic features from CLIP and identity features from ArcFace, project and fuse them into a unified identity representation, and process them via dual cross attention and a Fusion multilayer perceptron. The resulting embeddings are cross attended into a pre trained Stable Diffusion UNet to guide identity consistent denoising.}
    \label{fig:wide_figure}
\end{figure*}

\subsection{Model Architecture}
\textbf{Diff-ID} is built upon a pretrained Stable Diffusion 1.5 UNet architecture, chosen for its effectiveness in high resolution image synthesis within a compressed latent space~\cite{rombach2022high}. The architecture consists of several key modules designed to retain identity fidelity while maintaining the flexibility needed for morphing and prompt conditioned synthesis, without targeting explicit attribute editing.

\subsubsection{Embedding Extraction and Fusion}
In many prior models, identity features are extracted using a single embedding source, either via ArcFace (for example in Arc2Face) or through the CLIP image encoder (for example in IP Adapter and InstantID). However, relying solely on one modality can be limiting. ArcFace is highly effective at capturing fine grained, identity specific details that are crucial for face recognition, but it may not fully capture the semantic nuances necessary for aligning with textual descriptions. In contrast, the CLIP image encoder provides robust semantic representations that align well with text inputs, yet it may overlook subtle identity cues that are essential for preserving individual characteristics. To address these complementary strengths and weaknesses, Diff-ID integrates both embedding types, thereby ensuring a more comprehensive representation of identity and semantics.

\paragraph{CLIP Embeddings}  
Let \( A \) denote the input text prompt and \( I \) the reference image. The text embedding is computed as \( e_t = \text{CLIP}_{\text{text}}(A) \) while the image embedding is \( e_i = \text{CLIP}_{\text{image}}(I) \). Here, \(\text{CLIP}_{\text{text}}\) and \(\text{CLIP}_{\text{image}}\) represent the text and image encoding components of CLIP, respectively, with \( e_t \) encapsulating the semantic content of the prompt and \( e_i \) providing complementary visual context.

\paragraph{ArcFace Embeddings}  
Given the same input image \( I \), the ArcFace model produces a 512 dimensional identity embedding \( e_{\text{arc}} = \text{ArcFace}(I) \). This embedding captures unique identity specific features that are critical for maintaining identity fidelity during generation~\cite{deng2019arcface}.

\paragraph{Embedding Projection and Fusion}  
To integrate multi modal information, we first project the ArcFace and CLIP image embeddings into a common latent space. Their projected forms are given by
\begin{equation}
e_{\text{arc}}' = W_{\text{arc}}\, e_{\text{arc}},
\qquad
e_i' = W_i\, e_i,
\label{eq:embedding_projection}
\end{equation}
where \( W_{\text{arc}} \) and \( W_i \) are learnable projection matrices. Next, we concatenate these projected embeddings to form a combined image representation,
\begin{equation}
e_{\text{img}} = \text{Concat}(e_{\text{arc}}',\, e_i')\,.
\end{equation}
To further capture salient identity features, we apply both max pooling and average pooling on \( e_{\text{img}} \), yielding \( e_{\text{max}} = \text{MaxPool}(e_{\text{img}}) \) and \( e_{\text{avg}} = \text{AvgPool}(e_{\text{img}}) \). We then form the final identity representation by concatenating the individual components:
\begin{equation}
  \label{eq:id_embedding}
  e_{\mathrm{id}}
  = \mathrm{Concat}\bigl(e_{\mathrm{arc}}',\, e_i',\, e_{\max},\, e_{\mathrm{avg}}\bigr)\,.
\end{equation}
In parallel, the CLIP text embedding \( e_t \) is aligned with the identity modality by passing both \( e_{\text{id}} \) and \( e_t \) through a shared linear projection layer with weight matrix \( W_s \), resulting in 
\begin{equation}
  \label{eq:scaled_embeddings}
  \tilde{e}_{\mathrm{id}} = W_s\, e_{\mathrm{id}},
  \quad
  \tilde{e}_t = W_s\, e_t\,.
\end{equation}
This yields two separate yet aligned branches: the identity branch and the text branch for subsequent processing. Although \(\tilde{e}_{\mathrm{id}}\) and \(\tilde{e}_t\) share the same dimensionality, we deliberately avoid projecting ArcFace directly into the CLIP semantic space. ArcFace embeddings encode geometric identity under an angular margin objective, whereas CLIP encodes global semantic and contextual cues. Forcing them into a single shared representation collapses this complementarity and empirically degrades identity discriminability. Instead, Diff-ID maintains modality specific projections and fuses them only after cross attention refinement, preserving both fine grained identity cues and semantic alignment.

\subsubsection{Dual Cross Attention Mechanism}
To effectively integrate and refine the identity and textual information, we propose a dual cross attention mechanism. The motivation behind this design is twofold. First, by enabling bidirectional interactions, the model allows each modality to inform and enhance the other, ensuring that identity features are enriched by semantic cues from the prompt and conversely that semantic features are refined in an identity aware way. Second, by omitting softmax normalization, the mechanism preserves the raw magnitude relationships between features, which can help retain fine grained details that might otherwise be diminished. Importantly, we do not treat this as a dedicated attribute editing module: the goal is to stabilise identity under text conditioning, rather than to offer precise, controllable attribute manipulation.

We compute two cross attention branches jointly using the generic template
\begin{equation}
e = \left( \frac{Q K^{\top}}{\sqrt{d}} \right) V ,
\label{eq:cross_attention_template}
\end{equation}
where \( d \) is the embedding dimensionality.

\paragraph{Identity Branch}  
In the identity branch, the identity embedding \(\tilde{e}_{\text{id}}\) serves as the query and the CLIP text embedding \(\tilde{e}_t\) provides both the key and the value:
\begin{equation}
Q_{\text{id}} = W_q^{\text{id}}\, \tilde{e}_{\text{id}}, \quad
K_{\text{id}} = W_k^{\text{id}}\, \tilde{e}_t, \quad
V_{\text{id}} = W_v^{\text{id}}\, \tilde{e}_t.
\label{eq:cross_attention_id_branch}
\end{equation}
The enriched identity representation is then
\begin{equation}
e_{\text{identity}} = \left( \frac{Q_{\text{id}}\, K_{\text{id}}^{\top}}{\sqrt{d}} \right) V_{\text{id}}.
\label{eq:cross_attention_id_output}
\end{equation}

\paragraph{Text Branch}  
In the text branch, the CLIP text embedding \(\tilde{e}_t\) acts as the query, while the identity embedding \(\tilde{e}_{\text{id}}\) is used as both key and value:
\begin{equation}
Q_{\text{attr}} = W_q^{\text{attr}}\, \tilde{e}_t, \quad
K_{\text{attr}} = W_k^{\text{attr}}\, \tilde{e}_{\text{id}}, \quad
V_{\text{attr}} = W_v^{\text{attr}}\, \tilde{e}_{\text{id}},
\label{eq:cross_attention_attr_branch}
\end{equation}
\begin{equation}
e_{\text{attribute}} = \left( \frac{Q_{\text{attr}}\, K_{\text{attr}}^{\top}}{\sqrt{d}} \right) V_{\text{attr}}.
\label{eq:cross_attention_attr_output}
\end{equation}

\subsubsection{Fusion Multilayer Perceptron}
Although using a multilayer perceptron for feature fusion is standard practice, our design incorporates a Fusion multilayer perceptron module for an important reason: it enables the network to learn complex, nonlinear interactions between the enriched identity and text features. In doing so, we refine the joint representation in a way that preserves identity specific details while maintaining consistency with the textual description, without attempting explicit fine grained attribute control.

The outputs from the dual cross attention modules, the enriched identity representation \( e_{\text{identity}} \) and the enriched text representation \( e_{\text{attribute}} \), are concatenated to form a unified feature representation:
\begin{equation}
  \label{eq:fused_embedding}
  e_{\mathrm{fused}}
  = \mathrm{Concat}\bigl(e_{\mathrm{identity}},\, e_{\mathrm{attribute}}\bigr)\,.
\end{equation}
This fused embedding is then processed through a Fusion multilayer perceptron block to generate a refined embedding:
\begin{equation}
  \label{eq:refined_embedding}
  e_{\mathrm{refined}}
  = \mathrm{MLP}\bigl(e_{\mathrm{fused}}\bigr)\,,
\end{equation}
where the multilayer perceptron is defined as:
\begin{equation}
  \label{eq:mlp_definition}
  \mathrm{MLP}(x)
  = \sigma\!\bigl(W_{2}\,\sigma(W_{1} x + b_{1}) + b_{2}\bigr)\,.
\end{equation}
Here, \( W_1 \) and \( W_2 \) are learnable weight matrices, \( b_1 \) and \( b_2 \) are biases, and \(\sigma\) represents the rectified linear unit activation function.

\subsection{Objective Function}
\label{sec:objfun}
Our objective function is designed to balance two essential goals: high quality denoising and identity preservation. The denoising loss, adapted from the Stable Diffusion framework, drives the model to remove noise and reconstruct images that follow the target distribution. In parallel, the adversarial identity loss, computed using ArcFace embeddings, encourages generated images to retain core identity features. Recognizing that identity details are less discernible at higher noise levels, we incorporate an exponential dynamic weighting strategy. This strategy modulates the importance of identity loss throughout the diffusion process, emphasizing identity preservation when the noise is lower and identity information is more reliable.

\paragraph{Denoising Loss}  
Following the Stable Diffusion framework, the model minimizes the difference between the true noise \( \epsilon \) added during the forward process and the noise predicted by the network \( \epsilon_\theta(z_t, t) \). The denoising loss is given by:
\begin{equation}
\mathcal{L}_{\text{denoise}} = \mathbb{E}_{z_0, t, \epsilon} \left[ \| \epsilon - \epsilon_\theta(z_t, t) \|_2^2 \right]
\label{eq:loss_denoise}
\end{equation}
where \( t \) is uniformly sampled from \( \{1,2,\ldots,T\} \) and the noisy latent \( z_t \) is computed as:
\begin{equation}
z_t = \sqrt{\alpha_t}\,z_0 + \sqrt{1-\alpha_t}\,\epsilon, \quad \epsilon \sim \mathcal{N}(0,I).
\label{eq:latent_noising}
\end{equation}

\paragraph{Adversarial Identity Loss}  
To encourage generated images to preserve core identity features, ArcFace is employed as a pseudo discriminator. It extracts identity embeddings \( e_{\text{original}} \) and \( e_{\text{generated}} \) from the original and generated images, respectively. The identity loss is defined as
\begin{equation}
\mathcal{L}_{\text{identity}} = 1 - \text{CosineSim}(e_{\text{original}}, e_{\text{generated}})
\label{eq:identity_loss}
\end{equation}
where \(\text{CosineSim}\) denotes the cosine similarity between the two embeddings.

\paragraph{Exponential Dynamic Weighting Strategy}
Since identity features become less discernible at higher noise levels, we weight the identity loss exponentially based on the current timestep. Let \( \tau = \frac{t}{T} \) (with \( t \) as the current timestep and \( T \) as the total number of timesteps) and define the weight as
\begin{equation}
W_t = \exp\big(-k\,\tau\big),
\label{eq:tau_weight}
\end{equation}
with \( k \) as a decay hyperparameter controlling the rate of exponential decrease. The weighted identity loss per sample is computed as
\begin{equation}
\mathcal{L}_{\text{identity, weighted}} = \mathbb{E}\Big[W_t \cdot\mathcal{L}_{\text{identity}}\Big].
\label{eq:weighted_identity_loss}
\end{equation}
Intuitively, this schedule aligns the strength of identity supervision with the diffusion dynamics. At early timesteps the latent is dominated by Gaussian noise and identity cues are unreliable, so the identity loss is down weighted. At later timesteps, when the denoiser reconstructs semantically coherent faces, identity discrepancies become meaningful and the loss gradually plays a larger role.

\begin{figure}[ht]
  \centering
  \includegraphics[width=0.5\columnwidth]{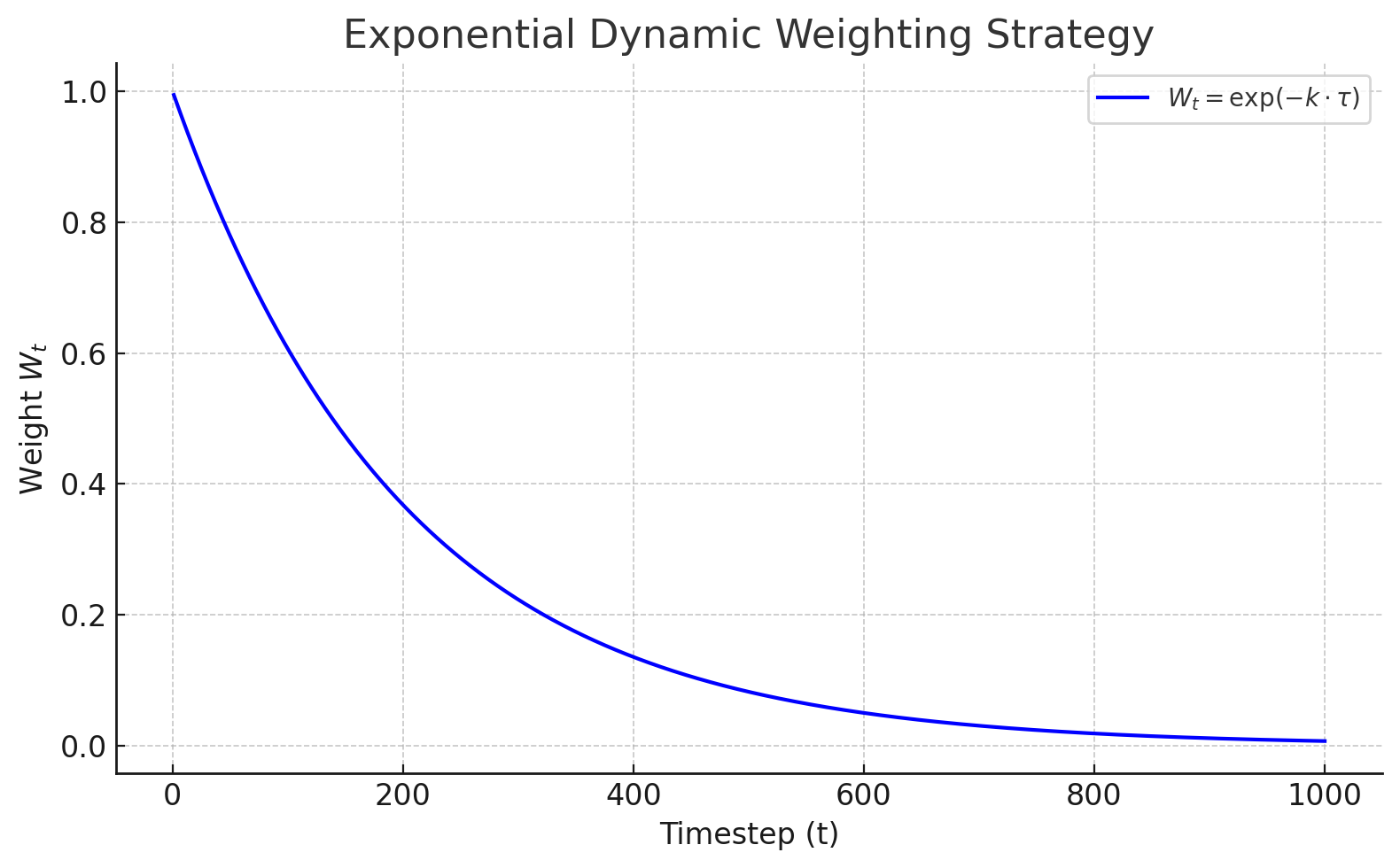}
  \caption{Illustration of the exponential timestep weighting \(W_t\) as a function of the normalized timestep \(\tau\). Identity supervision is weak at high noise levels and becomes increasingly important as the denoising process progresses.}
  \label{fig:identity_weight}
\end{figure}

\paragraph{Overall Training Objective}  
The final training loss is a combination of the denoising loss and the dynamically weighted identity loss:
\begin{equation}
\mathcal{L}_{\text{total}} = \mathcal{L}_{\text{denoise}} + \lambda_{\text{identity}} \cdot \mathcal{L}_{\text{identity, weighted}},
\label{eq:loss_total}
\end{equation}
where \( \lambda_{\text{identity}}=0.10 \) is a hyperparameter balancing the importance of identity preservation against denoising accuracy.

\subsection{Morphology Based Generation}
\begin{figure*}[ht]
    \centering
    \includegraphics[width=1.0\textwidth]{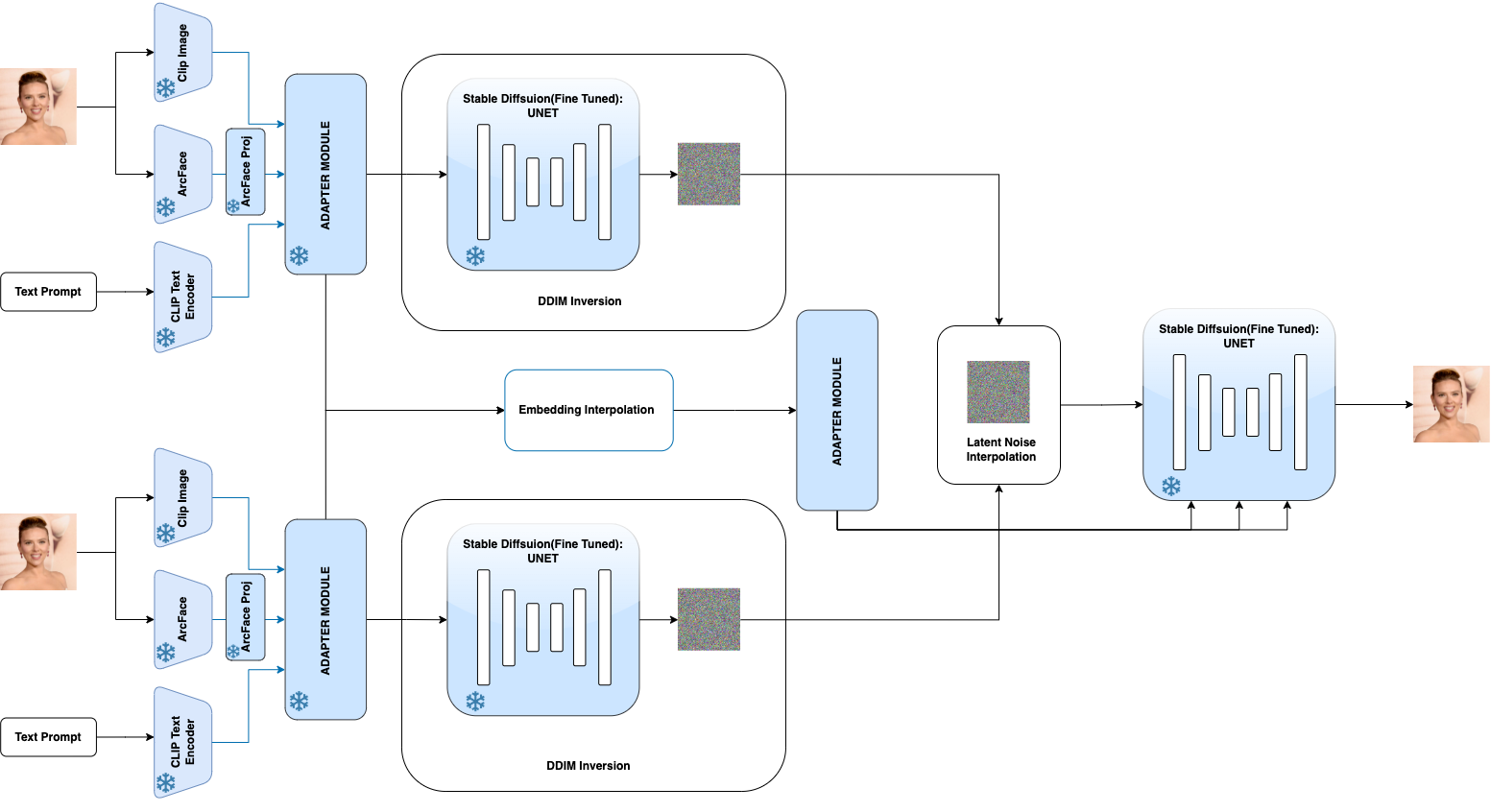}
    \caption{DiffID Morph: identity preserving morphing pipeline built on DDIM inversion and dual embedding interpolation.}
    \label{fig:DiffID-Morph}
\end{figure*}
We extend Diff-ID to support facial morphing between two source identities without requiring per identity fine tuning. Our method builds on the pretrained dual branch adapter and UNet backbone and leverages deterministic sampling via DDIM to produce stable qualitative trajectories. The generation proceeds in three key stages.

\paragraph{Stage 1: Inversion to Latent Space}  
Given two real input images \(x_1\) and \(x_2\), we use DDIM inversion~\cite{song2020denoising} to recover their corresponding noise latents \(z^T_1\) and \(z^T_2\) at the final timestep \(T\) of the diffusion process:
\begin{equation}
  \label{eq:ddim_inversion}
  z^{T} 
  = \mathrm{DDIMInvert}(x; \theta) 
  = \Phi^{-1}_{\theta}(x)\,,
\end{equation}
where \(\theta\) are the parameters of the pretrained denoising network and \(\Phi^{-1}_\theta\) represents the learned inversion mapping. This step aims to preserve both the visual and identity information of each input image in the diffusion space.

\paragraph{Stage 2: Embedding and Latent Interpolation}  
We extract identity embeddings \(e_1\) and \(e_2\) using the pretrained CLIP and ArcFace encoders, as integrated into the Diff-ID dual adapter module. To morph between the two identities, we perform spherical linear interpolation of the embeddings on the unit hypersphere:
\begin{equation}
  \label{eq:slerp}
  \begin{split}
    e_{\mathrm{mix}}
    &= \mathrm{SLERP}(e_1, e_2; \alpha) \\
    &= \frac{\sin\bigl((1-\alpha)\,\omega\bigr)}{\sin\omega}\,e_1
     + \frac{\sin\bigl(\alpha\,\omega\bigr)}{\sin\omega}\,e_2, \\
    \omega
    &= \arccos\!\Bigl(\frac{e_1 \cdot e_2}{\|e_1\|\|e_2\|}\Bigr),
    \quad \alpha \in [0,1]
  \end{split}
\end{equation}
where \(\alpha\) controls the morphing intensity between the two faces. In parallel, we interpolate the latent noise vectors \(z^T_1\) and \(z^T_2\) using linear interpolation:
\begin{equation}
  \label{eq:mix_noise}
  z^{T}_{\mathrm{mix}}
  = (1 - \alpha)\,z^{T}_{1}
  + \alpha\,z^{T}_{2}\,,
\end{equation}
which encourages the structure and fine details of both identities to be blended in the denoising trajectory. Combining both embedding and latent interpolation is intended to support smooth transitions in both semantic identity space and low level generative signal.

\paragraph{Stage 3: Identity Conditioned Sampling}  
Using the interpolated noise vector \(z^T_{\mathrm{mix}}\) and the blended identity embedding \(e_{\mathrm{mix}}\), we perform deterministic DDIM sampling guided by the Diff-ID dual cross attention mechanism:
\begin{equation}
  \label{eq:ddim_sampling}
  \hat{x}
  = \mathrm{DDIMSample}\bigl(z^{T}_{\mathrm{mix}};\,e_{\mathrm{mix}},\,\theta\bigr)\,,
\end{equation}
producing the final morphed image \(\hat{x}\) with characteristics from both source identities and the goal of maintaining realism and coherence. Our approach supports continuous control over the morph factor \(\alpha\), which enables a spectrum of identities between \(x_1\) and \(x_2\). 

Unlike other methods that require multiple fine tuned checkpoints, for example DreamBooth~\cite{ru2022dreambooth} or DiffMorpher~\cite{lee2024diffmorpher}, or low rank adaptation layers such as low rank adaptation based fine tuning~\cite{hu2021lora}, our pipeline operates in a single unified model. This avoids per identity checkpoint swapping and may support future applications such as identity obfuscation~\cite{rathgeb2019survey}, biometric robustness testing~\cite{ferrara2014magic}, and photorealistic avatar creation in augmented reality and virtual reality systems via three dimensional morphable models~\cite{blanz1999morphable}, pending task specific evaluation.

\subsection{Training Configuration and Hyperparameters}
The training setup for Diff-ID involves a mix of frozen and trainable layers, with the following components kept frozen to retain their pretrained feature extraction capabilities:
\begin{itemize}
    \item \textbf{Variational Autoencoder Layers}: Encoder and decoder layers of the variational autoencoder are frozen to maintain the latent to image space transformations without additional fine tuning~\cite{rombach2022high}.
    \item \textbf{CLIP and ArcFace Models}: Both models are kept frozen, preserving their pretrained strengths in capturing semantic (CLIP) and identity specific (ArcFace) features~\cite{deng2019arcface}.
\end{itemize}

Diff-ID was trained using the Adam optimizer with a learning rate of \(1 \times 10^{-5}\) and a batch size of 16, employing mixed precision (fp16) for efficiency. Gradient checkpointing was utilized to manage memory consumption, which allows larger batch sizes without exceeding graphics processing unit memory limits. The training was conducted over 1{,}000{,}000 steps using four NVIDIA A100 80GB graphics processing units, achieving convergence and stable performance in approximately 72 hours.

% ----------------------------------------------------------------------
% 4. Evaluation and Results
% ----------------------------------------------------------------------

\section{Evaluation and Results}
\label{Evaluation_and_Results}

\subsection{Evaluation Framework}
To thoroughly assess Diff-ID effectiveness, we employ a two pronged evaluation strategy that combines qualitative inspection with quantitative metrics targeting identity preservation and visual realism.

\subsubsection{Qualitative Analysis}  
Following established practice~\cite{liu2023ip, zhang2023controlnet}, we present side by side image grids comparing Diff-ID outputs with those of baseline methods. This visual inspection emphasizes:
\begin{itemize}
  \item \textbf{Identity Fidelity:} The extent to which characteristic facial features are retained.  
  \item \textbf{Perceptual Realism:} The absence of unnatural artifacts or distortions.  
\end{itemize}
Because all models are driven by the same BLIP generated descriptive prompts for each identity, the qualitative comparisons also reveal whether semantic attributes (for example hair colour, apparent age cues, coarse expression, and lighting) remain consistent with the caption. Diff-ID is not designed as an explicit attribute editing system, and we do not claim or evaluate fine grained attribute control. In practice, the reciprocal cross attention in Section~\ref{sec:methodology} can still yield some identity preserving, prompt aligned variation in these high level attributes while maintaining geometry and identity defining structure, but this behaviour is incidental rather than a primary objective.

\subsubsection{Quantitative Analysis}  
We evaluate two complementary dimensions of performance.

\paragraph{Identity Preservation}  
To quantify how well a generated image maintains the identity of the target subject, we use the \textbf{Face Similarity (FS)} metric. FS is computed as the cosine similarity between the 512-dimensional embeddings extracted by a pretrained ArcFace model~\cite{deng2019arcface} for each source--generated image pair. Since ArcFace is optimized for face recognition, FS provides a strong and widely adopted proxy for measuring identity fidelity. However, we note that FS measures identity independently of visual realism or style, and may therefore assign high similarity scores to highly stylized, painted, or otherwise non-photorealistic outputs, provided that the underlying facial geometry is preserved. This behaviour is illustrated in Figure~\ref{fig:variants_fs}, where stylistically diverse images still yield high FS values.

\paragraph{Visual Realism}  
We measure realism using the Fréchet Inception Distance~\cite{heusel2017gans}, which quantifies the statistical divergence between the distributions of generated and real images in the Inception v3 feature space. While the Fréchet Inception Distance captures large scale discrepancies, it can sometimes under represent small, but perceptually significant, facial artifacts.

\subsubsection{Face Image Quality Composite Score}

\begin{figure}[ht]
  \centering
  \setlength{\tabcolsep}{2pt}
  \renewcommand{\arraystretch}{1.0}
  \begin{tabular}{ccccc}
    \textbf{Original} & \textbf{Variant 1} & \textbf{Variant 2} & \textbf{Variant 3} & \textbf{Variant 4} \\

    \includegraphics[height=2.5cm]{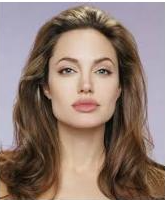} &
    \includegraphics[height=2.5cm]{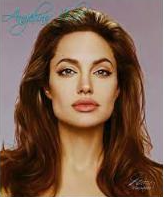} &
    \includegraphics[height=2.5cm]{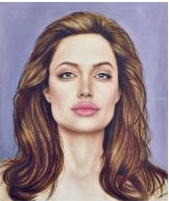} &
    \includegraphics[height=2.5cm]{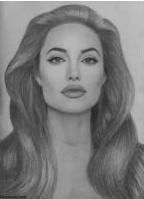} &
    \includegraphics[height=2.5cm]{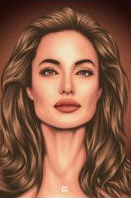} \\[4pt]

    \textbf{FS\%} & 89.55 & 68.44 & 83.71 & 61.55 \\
  \end{tabular}
  \caption{Face Similarity (FS) for one subject under four stylistic variations. FS remains high even for non-photorealistic or stylized outputs, provided facial structure and geometry are preserved.}
  \label{fig:variants_fs}
\end{figure}

Identity preservation in generative face models is commonly quantified using Face Similarity (FS), typically computed using embedding-based cosine similarity from recognition models such as ArcFace. FS is expressed on a \([0,100]\) scale and measures how well the generated face maintains the target identity. However, FS evaluates identity independently of realism or style, meaning that it may assign high similarity scores to images that are heavily stylized, artistic, or non-photorealistic, as long as they preserve the underlying geometric features of the face.

Figure~\ref{fig:variants_fs} demonstrates this behaviour. Despite significant stylistic variation across the four generated images, FS remains consistently high because the essential identity-defining structure is retained. This highlights an important limitation: while FS is effective for identity comparison, it does not penalize unrealistic textures, artefacts, or deviations from natural image statistics. Consequently, relying solely on FS can be misleading when evaluating the quality of generative models, as it cannot distinguish between realistic and unrealistic renditions of a face.

To address this, realism is typically assessed using the Fréchet Inception Distance (FID), which measures distributional similarity between generated images and real face distributions. While FID captures realism effectively, it ignores identity preservation. Thus, FS and FID each measure complementary but independent aspects of generative performance.

As a complementary summary of both identity fidelity and perceptual realism, we report the \textbf{Face Image Quality (FIQ)} score.

\begin{figure}[ht]
  \centering
  \setlength{\tabcolsep}{2pt}
  \renewcommand{\arraystretch}{1.0}
  \begin{tabular}{cccccc}
    \textbf{Orig.} & \textbf{IPAdpt} & \textbf{Photo} & \textbf{InstID} & \textbf{Arc2F} & \textbf{Ours} \\

    \includegraphics[height=2.2cm]{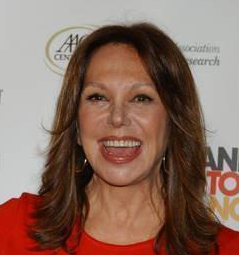} &
    \includegraphics[height=2.2cm]{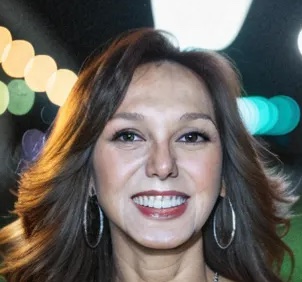} &
    \includegraphics[height=2.2cm]{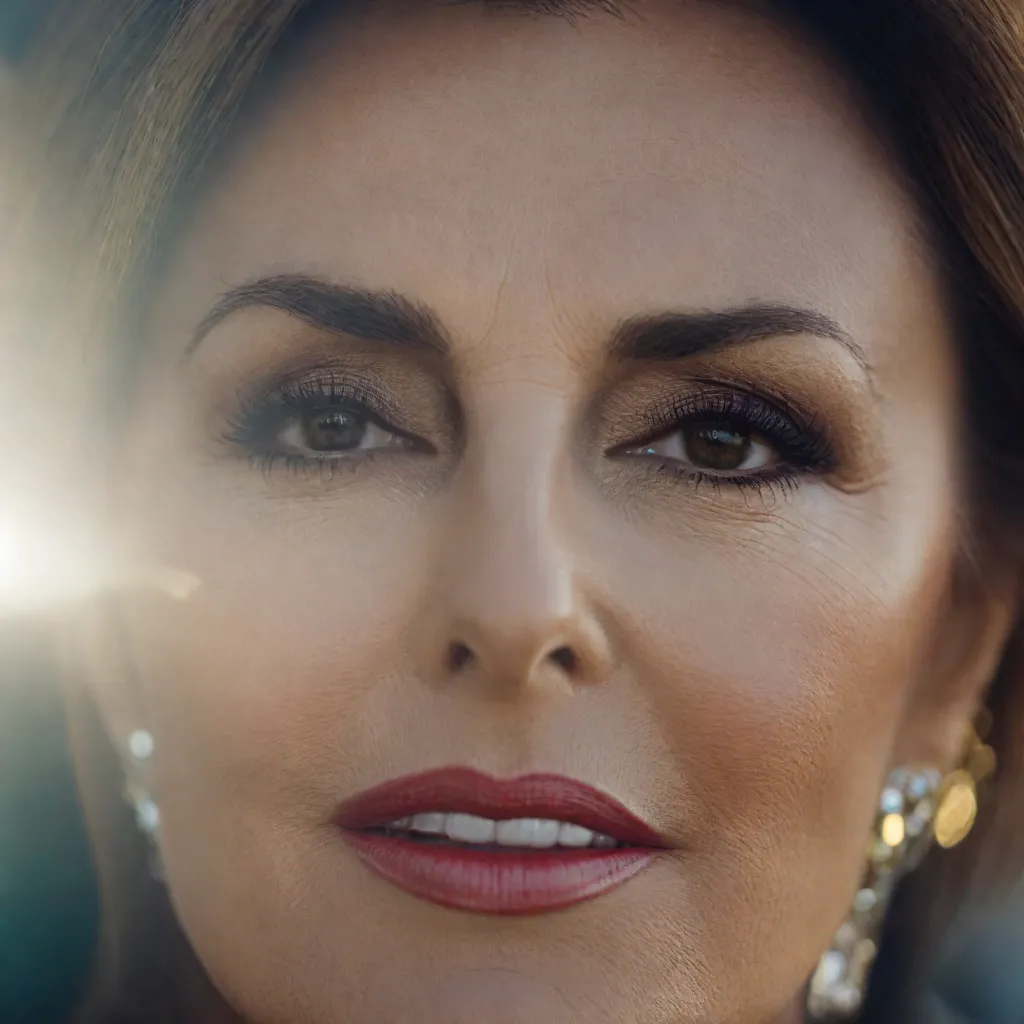} &
    \includegraphics[height=2.2cm]{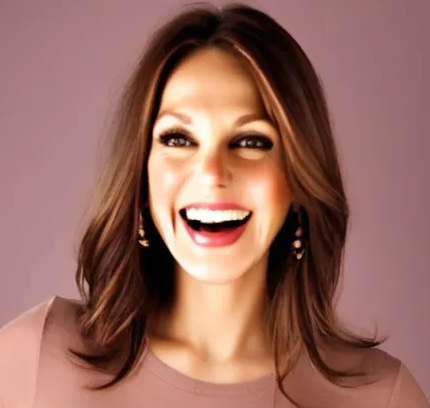} &
    \includegraphics[height=2.2cm]{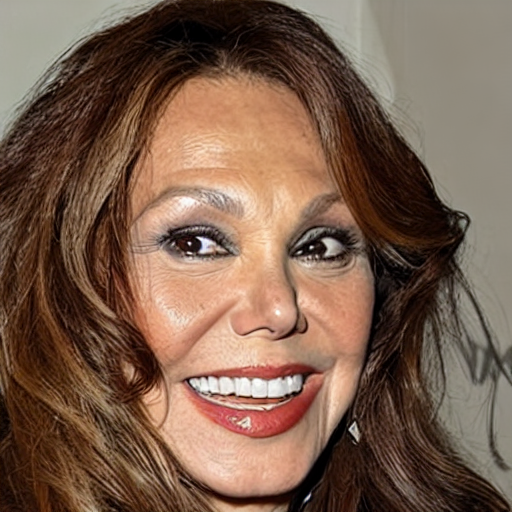} &
    \includegraphics[height=2.2cm]{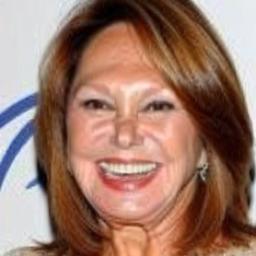} \\[2pt]

    \textbf{FS $\uparrow$}  & 47.92 & 0.09 & 42.82 & 63.15 & 81.42 \\
    \textbf{FID $\downarrow$} & 171.7 & 127.5 & 129.9 & 136.3 & 101.3 \\
    \textbf{FIQ $\uparrow$} & 27.91 & 0.07 & 32.96 & 46.33 & 80.38 \\

  \end{tabular}
  \caption{Identity-specific comparison for one subject (ID1) across different models. Metrics (FS, FID, and FIQ) are shown below each result.}
  \label{fig:id1_half_column}
\end{figure}

To summarize identity preservation and realism in a single secondary score, we define FIQ as the normalized ratio of FS to FID:

\begin{equation}
\label{eq:fiq}
\mathrm{FIQ} = 100 \cdot \frac{\mathrm{FS}}{\mathrm{FID}}\,.
\end{equation}

This formulation provides a simple coupled dependency between identity and realism. Higher FS increases FIQ, reflecting better identity alignment, while higher FID (indicating poorer realism) decreases FIQ, penalizing degraded or unrealistic outputs. Although the equation includes a factor of 100, FIQ is an unbounded ratio and should not be interpreted as a percentage.

We therefore treat FIQ as a diagnostic and comparative aid rather than a replacement for FS or FID. Since we do not validate FIQ against human judgments or biometric decision outcomes in this paper, the primary evidence remains the separate FS and FID values, with FIQ used to summarize their trade off.

\subsection{Evaluation Settings}
For quantitative evaluation, we use two image sets of size 5{,}000 each. The first is a validation split of 5{,}000 images sampled from our curated CelebA-HQ, FFHQ, and LAION-Face pool, each paired with a descriptive caption produced by the BLIP based captioning system described in Section~\ref{sec:methodology}. The second is an unseen split of 5{,}000 images drawn from the LFW dataset, again captioned using the same BLIP model to obtain identity aware prompts. All methods, including Diff-ID and the baselines in Table~\ref{tab:evaluation}, are conditioned on these captions when generating outputs, providing a consistent comparison across identity fidelity and realism on both validation and unseen identities.

Identity disjointness between the training pool and the evaluation splits was not inherently enforced through a dedicated identity-matching audit; rather, it was assumed at the dataset level, on the premise that different source datasets are unlikely to share identities, though imprecise or incidental identity overlap cannot be ruled out. The validation split is sampled from the same curated CelebA-HQ, FFHQ, and LAION-Face pool used for training and may therefore overlap with the training identity distribution, so it should be treated as an in-domain rather than an identity-disjoint split. The unseen split, drawn from LFW, is not part of the Diff-ID training pool; as far as we are aware we aimed to keep this split identity-disjoint from training, but we treat this as a best-effort assumption rather than a formally verified guarantee.

\subsection{Results}

\subsubsection{Identity Preservation}
\begin{table}[!htbp]
  \centering
  \small
  \setlength{\tabcolsep}{3pt}
  \renewcommand{\arraystretch}{1.0}
  \begin{tabular}{l ccc ccc}
    \hline
    & \multicolumn{3}{c}{Validation Set} & \multicolumn{3}{c}{Unseen Data} \\
    \cline{2-4} \cline{5-7}
    Model 
    & FS $\uparrow$ & FID $\downarrow$ & FIQ $\uparrow$
    & FS $\uparrow$ & FID $\downarrow$ & FIQ $\uparrow$ \\
    \hline
    IP Adapter & 40.53  & 171.71  & 23.60 & 35.90  & 151.56  & 23.69 \\
    PhotoMaker & 33.87  & 127.47  & 26.57 & 29.56  & 113.50  & 26.04 \\
    InstantID  & \textbf{75.13} & 129.98 & 57.80 & \textbf{74.12} & 119.39 & 62.08 \\
    Arc2Face   & 73.71  & 136.25  & 54.10 & 72.51  & 110.69  & 65.51 \\
    Diff-ID    & 72.68  & \textbf{101.31} & \textbf{71.74}
               & 71.53  & \textbf{103.19} & \textbf{69.32} \\
    \hline
  \end{tabular}
  \caption{Evaluation: Face Similarity (FS), Fréchet Inception Distance (FID), and Face Image Quality (FIQ) on validation and unseen sets. Arrows indicate whether higher ($\uparrow$) or lower ($\downarrow$) values are better.}
  \label{tab:evaluation}
\end{table}

Because any single composite score can be sensitive to its functional form, we additionally check whether the measured identity--realism conclusion is stable under alternative formulations. Besides the original ratio based score \(\mathrm{FIQ}_{R}=100\cdot\mathrm{FS}/\mathrm{FID}\), we compute smoothed ratios \(\mathrm{FIQ}_{\alpha}=100\cdot\mathrm{FS}/(\mathrm{FID}+\alpha)\) for \(\alpha\in\{10,20,30,50\}\). To address the unbounded nature of the original ratio, we also define normalized identity and realism terms
\[
I=\frac{\mathrm{FS}}{100},
\qquad
R=\frac{1}{1+\mathrm{FID}/100},
\]
where 100 is used as a simple scale factor matching the order of magnitude of FID values in our evaluation; this normalization is used only for sensitivity analysis and not as a primary metric. We then report bounded harmonic and geometric composites
\[
\mathrm{FIQ}_{H}=\frac{2IR}{I+R},
\qquad
\mathrm{FIQ}_{G}=\sqrt{IR}.
\]
These bounded scores have clear limiting behaviour: if identity similarity collapses then \(I=0\), if realism degrades severely then \(R\rightarrow0\), and if both identity and realism are ideal then \(I=R=1\). These variants are not intended to replace FS, FID, or the original FIQ ratio; they are used only as a sensitivity analysis for the trade off conclusion. As shown in Table~\ref{tab:fiq_robustness}, Diff-ID remains highest under all tested composite formulations on both splits, while InstantID retains the highest raw Face Similarity. This suggests that Diff-ID's measured identity--realism advantage is not an artifact of the particular original ratio formulation.

\begin{table}[ht]
\centering
\begingroup
\scriptsize
\setlength{\tabcolsep}{3pt}
\renewcommand{\arraystretch}{1.08}
\resizebox{\linewidth}{!}{%
\begin{tabular}{lccccccccc}
\multicolumn{10}{c}{\textbf{Validation split}}\\
\hline
\textbf{Method} & \textbf{FS}\(\uparrow\) & \textbf{FID}\(\downarrow\) & \(\mathbf{FIQ_R}\)\(\uparrow\) & \(\mathbf{FIQ_{10}}\)\(\uparrow\) & \(\mathbf{FIQ_{20}}\)\(\uparrow\) & \(\mathbf{FIQ_{30}}\)\(\uparrow\) & \(\mathbf{FIQ_{50}}\)\(\uparrow\) & \(\mathbf{FIQ_H}\)\(\uparrow\) & \(\mathbf{FIQ_G}\)\(\uparrow\)\\
\hline
IP Adapter & 40.53 & 171.71 & 23.60 & 22.30 & 21.14 & 20.09 & 18.28 & 0.386 & 0.386\\
PhotoMaker & 33.87 & 127.47 & 26.57 & 24.64 & 22.97 & 21.51 & 19.08 & 0.383 & 0.386\\
InstantID & \textbf{75.13} & 129.98 & 57.80 & 53.67 & 50.09 & 46.96 & 41.74 & 0.551 & 0.572\\
Arc2Face & 73.71 & 136.25 & 54.10 & 50.40 & 47.17 & 44.34 & 39.58 & 0.538 & 0.559\\
Diff-ID & 72.68 & \textbf{101.31} & \textbf{71.74} & \textbf{65.30} & \textbf{59.91} & \textbf{55.35} & \textbf{48.03} & \textbf{0.590} & \textbf{0.601}\\
\hline
\multicolumn{10}{c}{}\\[-2pt]
\multicolumn{10}{c}{\textbf{Unseen split}}\\
\hline
\textbf{Method} & \textbf{FS}\(\uparrow\) & \textbf{FID}\(\downarrow\) & \(\mathbf{FIQ_R}\)\(\uparrow\) & \(\mathbf{FIQ_{10}}\)\(\uparrow\) & \(\mathbf{FIQ_{20}}\)\(\uparrow\) & \(\mathbf{FIQ_{30}}\)\(\uparrow\) & \(\mathbf{FIQ_{50}}\)\(\uparrow\) & \(\mathbf{FIQ_H}\)\(\uparrow\) & \(\mathbf{FIQ_G}\)\(\uparrow\)\\
\hline
IP Adapter & 35.90 & 151.56 & 23.69 & 22.22 & 20.93 & 19.77 & 17.81 & 0.377 & 0.378\\
PhotoMaker & 29.56 & 113.50 & 26.04 & 23.94 & 22.14 & 20.60 & 18.08 & 0.362 & 0.372\\
InstantID & \textbf{74.12} & 119.39 & 62.08 & 57.28 & 53.17 & 49.62 & 43.76 & 0.564 & 0.581\\
Arc2Face & 72.51 & 110.69 & 65.51 & 60.08 & 55.48 & 51.54 & 45.12 & 0.574 & 0.587\\
Diff-ID & 71.53 & \textbf{103.19} & \textbf{69.32} & \textbf{63.19} & \textbf{58.06} & \textbf{53.71} & \textbf{46.69} & \textbf{0.583} & \textbf{0.593}\\
\hline
\end{tabular}%
}
\caption{Robustness of identity--realism composite scoring. FS and FID remain the primary metrics. Alternative composite scores are reported only to test whether the identity--realism conclusion depends on the original FIQ ratio. Across both validation and unseen splits, Diff-ID remains highest under all tested composite formulations, while InstantID retains the highest raw FS and Diff-ID retains the lowest FID.}
\label{tab:fiq_robustness}
\endgroup
\end{table}

\begin{figure}[!htbp]
  \centering
  \setlength{\tabcolsep}{2pt}
  \begin{tabular}{ccccc}
    \textbf{Original} &
    \includegraphics[width=0.16\columnwidth,height=2.6cm,keepaspectratio]{ID1.png} &
    \includegraphics[width=0.16\columnwidth,height=2.6cm,keepaspectratio]{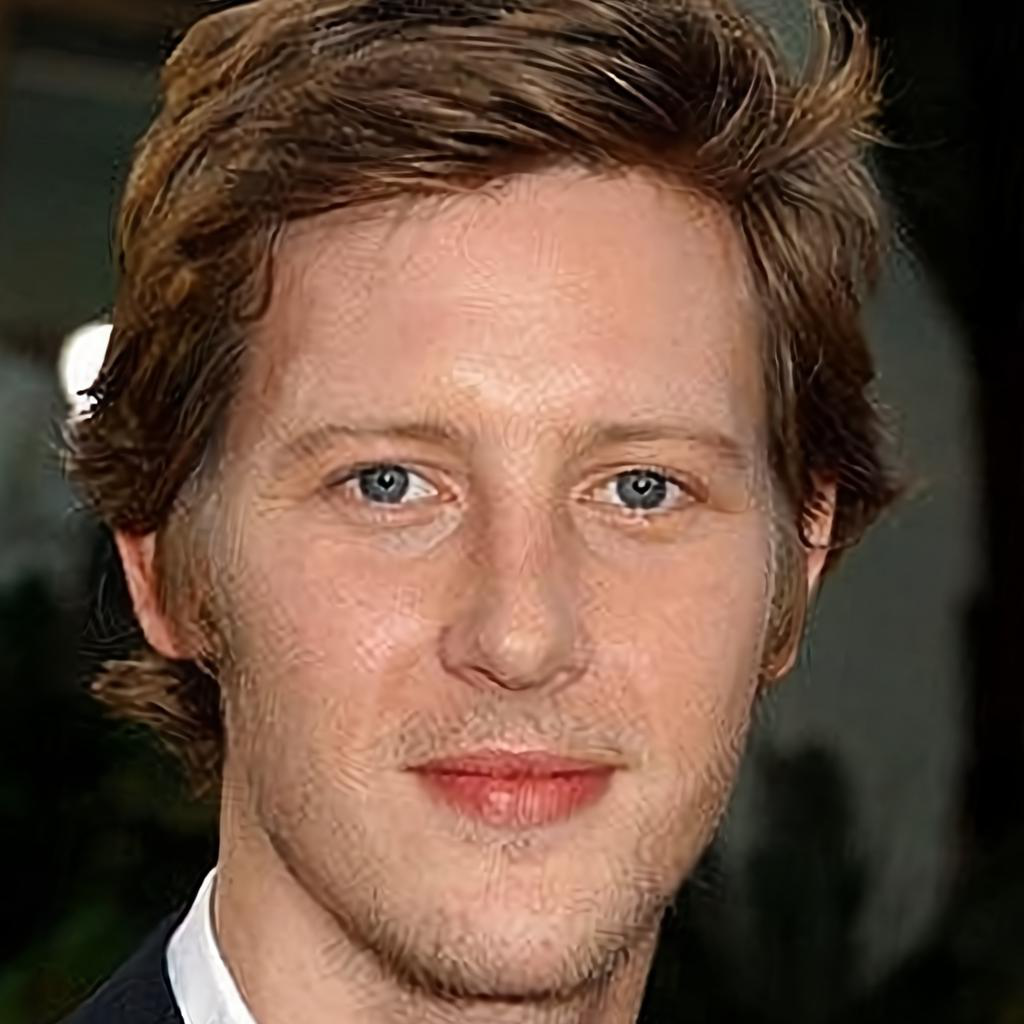} &
    \includegraphics[width=0.16\columnwidth,height=2.6cm,keepaspectratio]{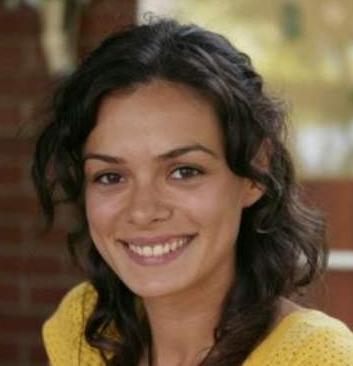} &
    \includegraphics[width=0.16\columnwidth,height=2.6cm,keepaspectratio]{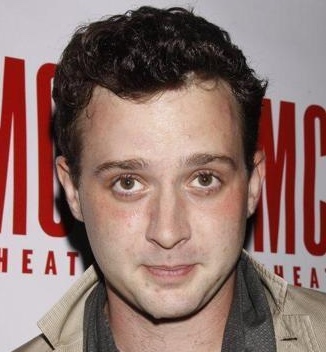} \\[2pt]

    \textbf{IP Adapter} &
    \includegraphics[width=0.16\columnwidth,height=2.6cm,keepaspectratio]{ID1IPadpater.jpg} &
    \includegraphics[width=0.16\columnwidth,height=2.6cm,keepaspectratio]{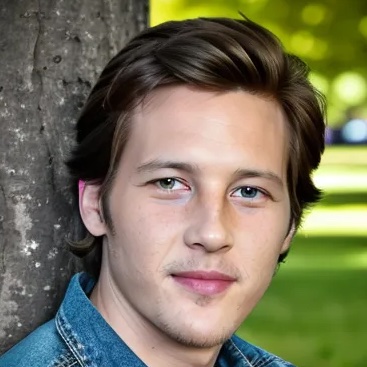} &
    \includegraphics[width=0.16\columnwidth,height=2.6cm,keepaspectratio]{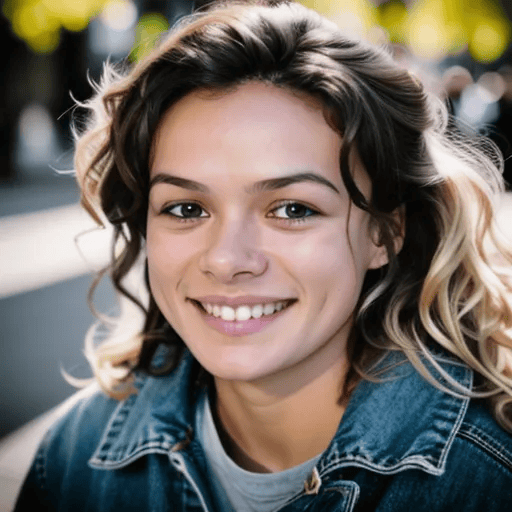} &
    \includegraphics[width=0.16\columnwidth,height=2.6cm,keepaspectratio]{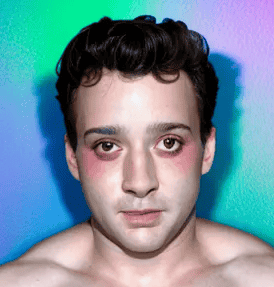} \\[2pt]

    \textbf{PhotoMaker} &
    \includegraphics[width=0.16\columnwidth,height=2.6cm,keepaspectratio]{ID1Photomaker.png} &
    \includegraphics[width=0.16\columnwidth,height=2.6cm,keepaspectratio]{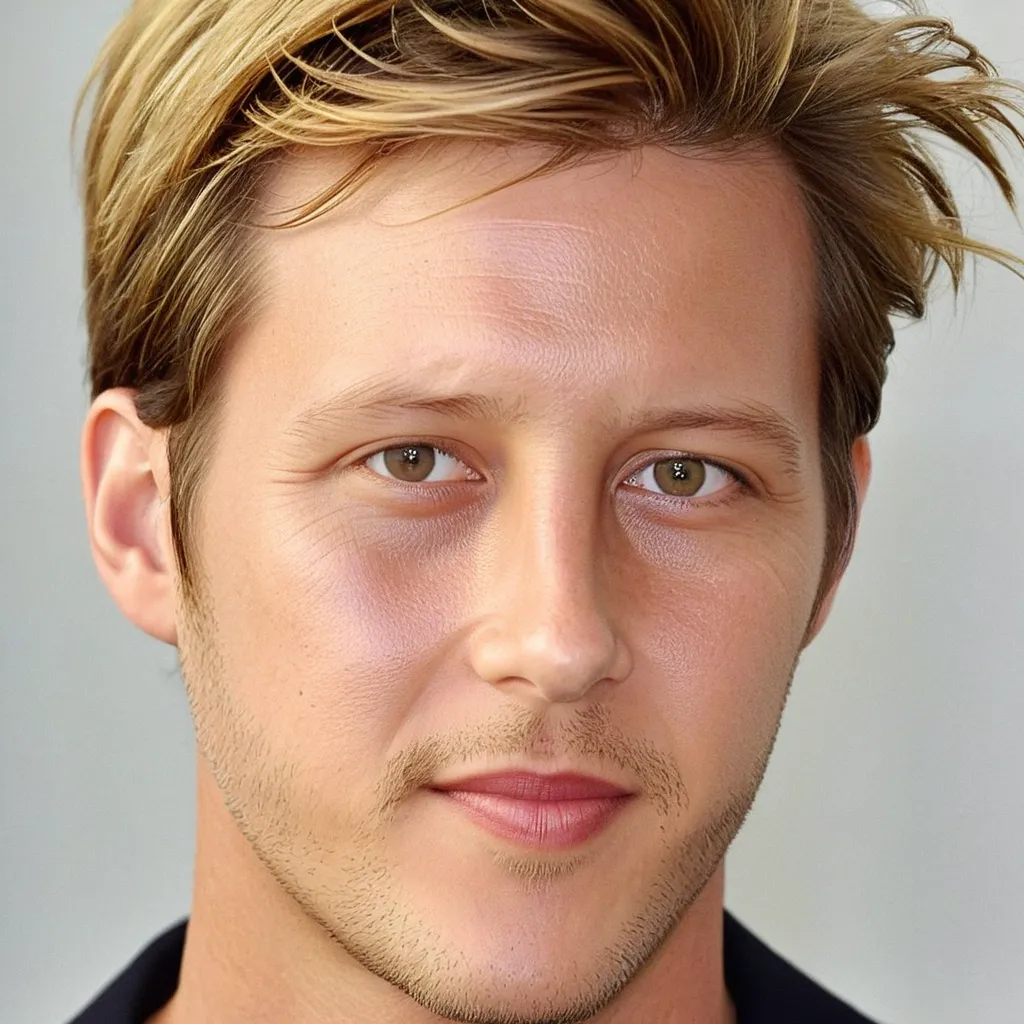} &
    \includegraphics[width=0.16\columnwidth,height=2.6cm,keepaspectratio]{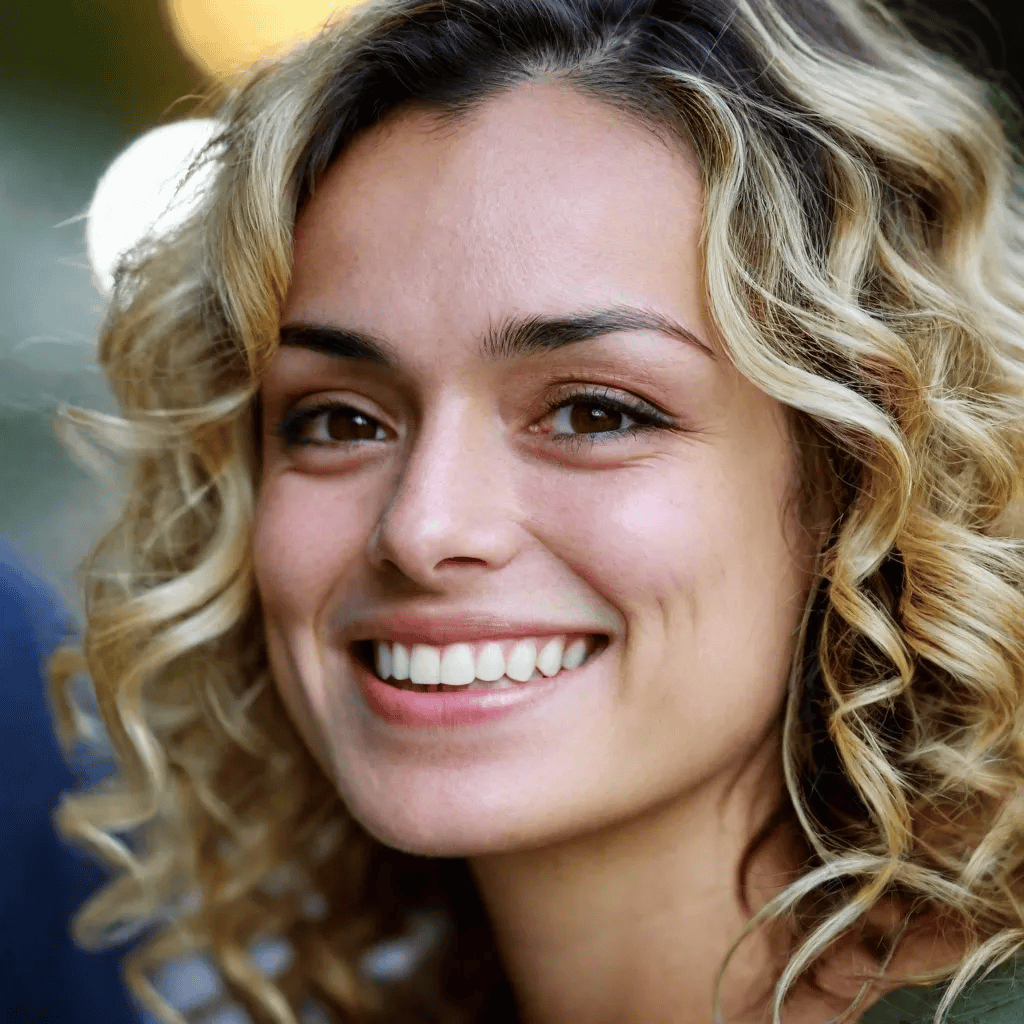} &
    \includegraphics[width=0.16\columnwidth,height=2.6cm,keepaspectratio]{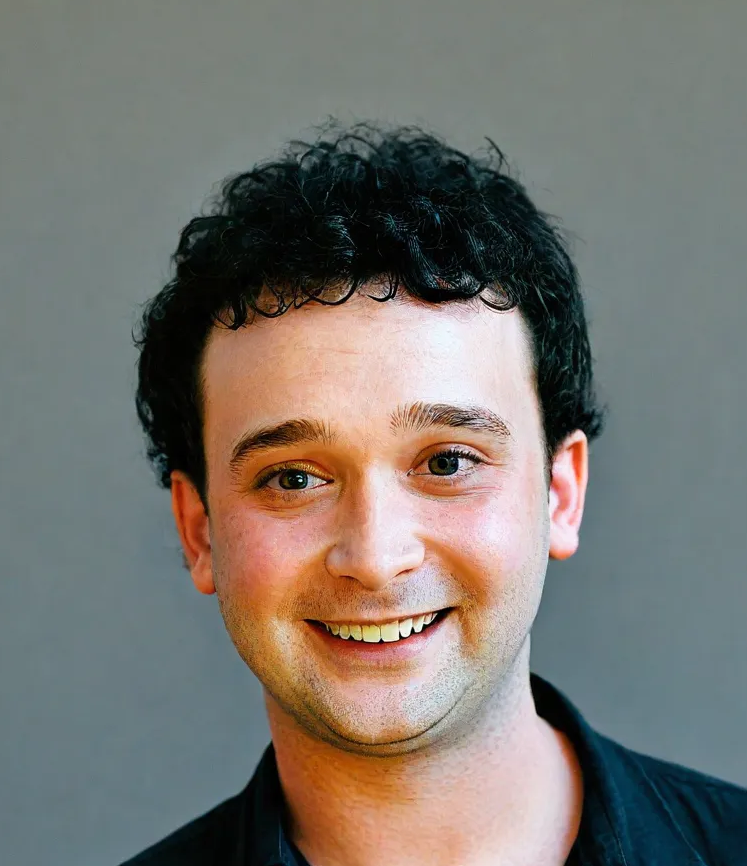} \\[2pt]

    \textbf{InstantID} &
    \includegraphics[width=0.16\columnwidth,height=2.6cm,keepaspectratio]{ID1Instantid.jpg} &
    \includegraphics[width=0.16\columnwidth,height=2.6cm,keepaspectratio]{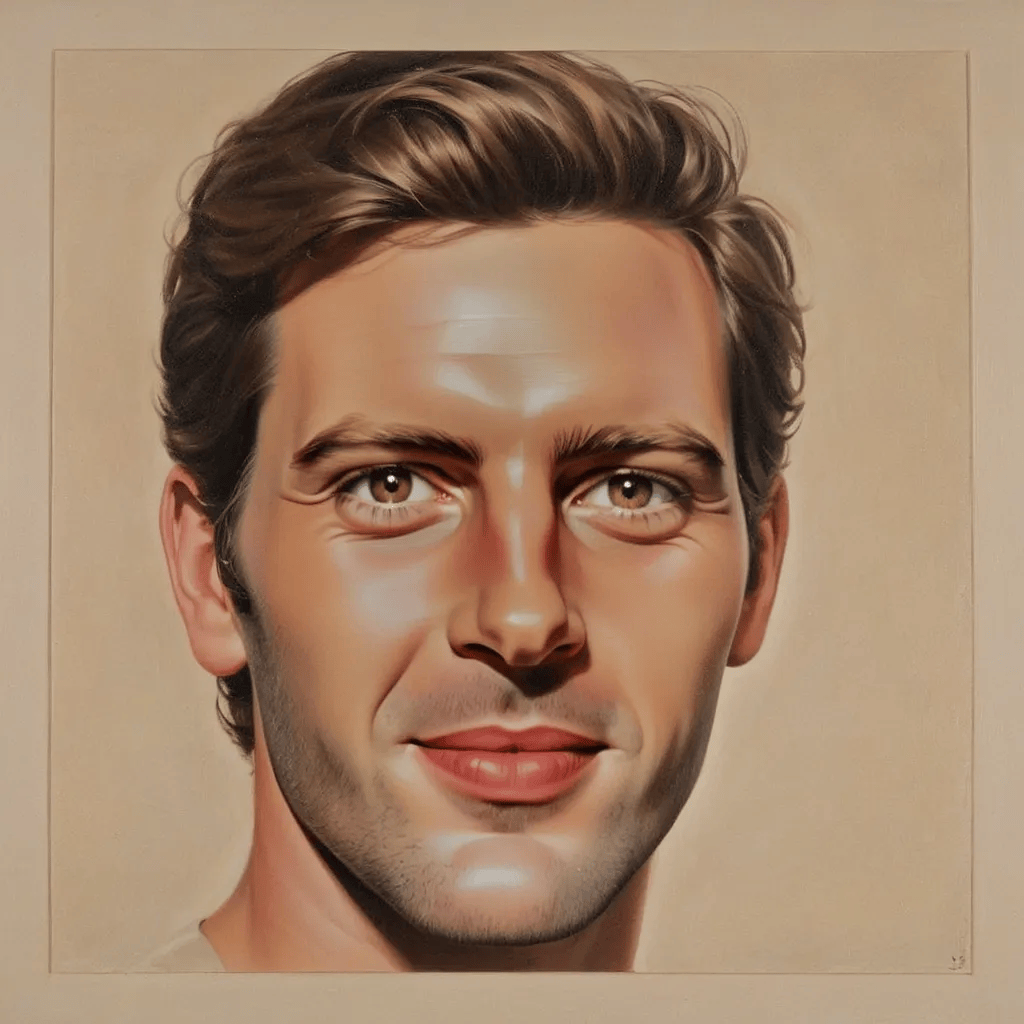} &
    \includegraphics[width=0.16\columnwidth,height=2.6cm,keepaspectratio]{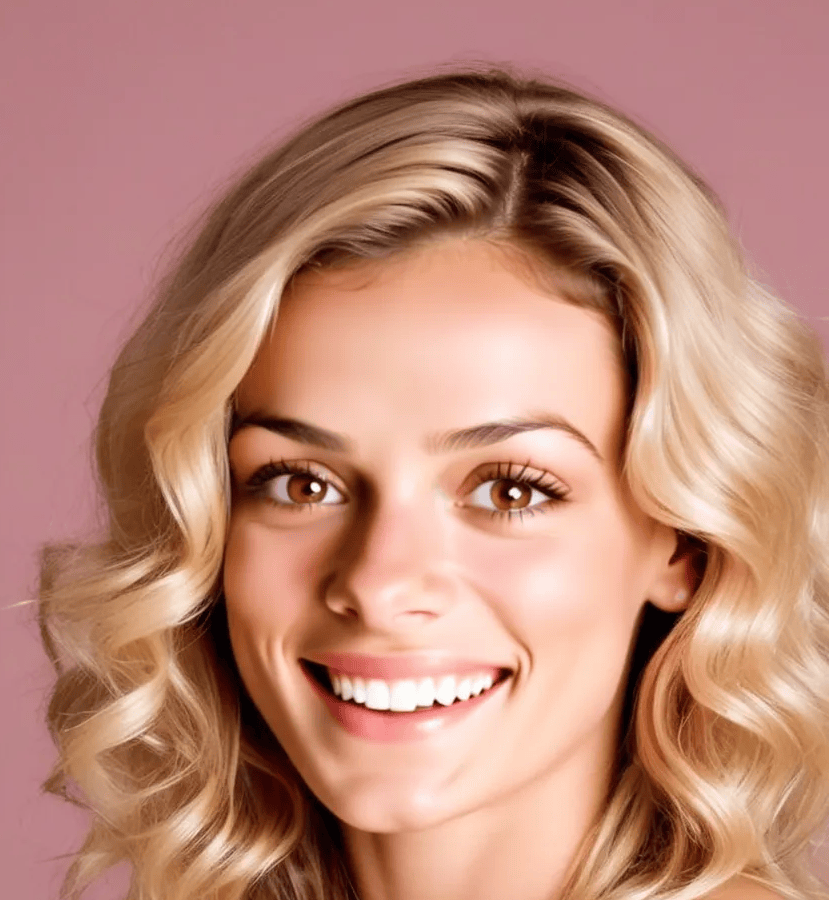} &
    \includegraphics[width=0.16\columnwidth,height=2.6cm,keepaspectratio]{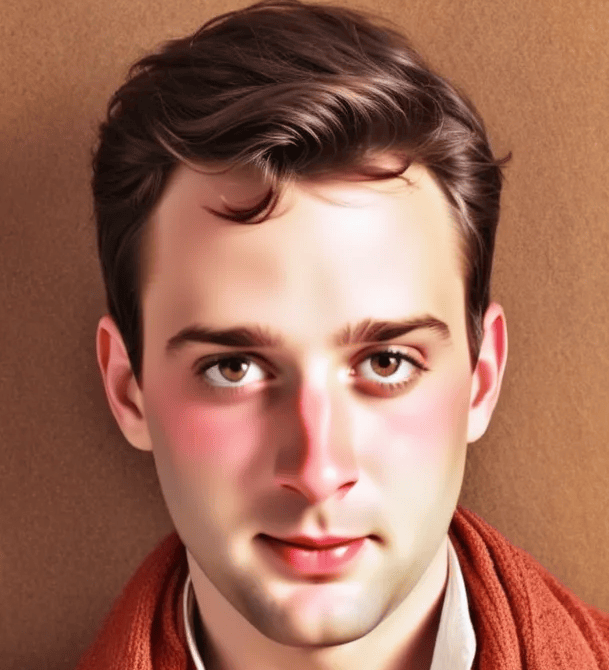} \\[2pt]

    \textbf{Arc2Face} &
    \includegraphics[width=0.16\columnwidth,height=2.6cm,keepaspectratio]{ID1Arc2face.png} &
    \includegraphics[width=0.16\columnwidth,height=2.6cm,keepaspectratio]{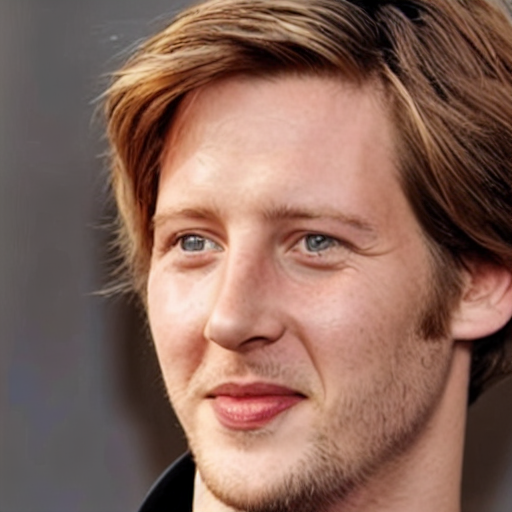} &
    \includegraphics[width=0.16\columnwidth,height=2.6cm,keepaspectratio]{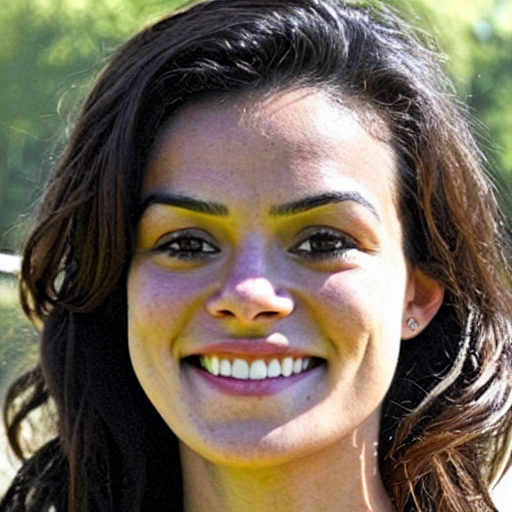} &
    \includegraphics[width=0.16\columnwidth,height=2.6cm,keepaspectratio]{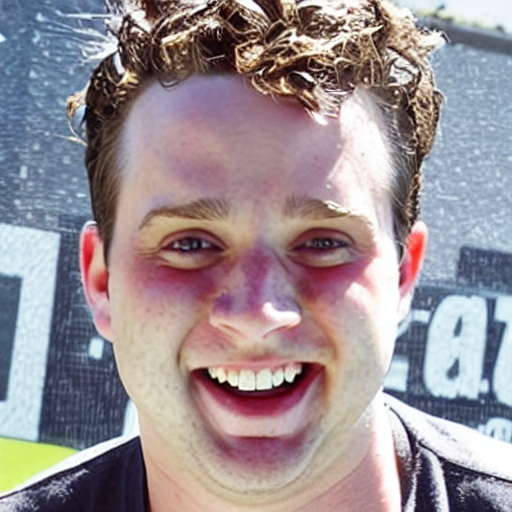} \\[2pt]

    \textbf{Ours} &
    \includegraphics[width=0.16\columnwidth,height=2.6cm,keepaspectratio]{DiffidID1.jpeg} &
    \includegraphics[width=0.16\columnwidth,height=2.6cm,keepaspectratio]{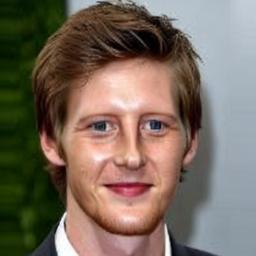} &
    \includegraphics[width=0.16\columnwidth,height=2.6cm,keepaspectratio]{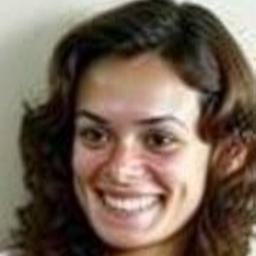} &
    \includegraphics[width=0.16\columnwidth,height=2.6cm,keepaspectratio]{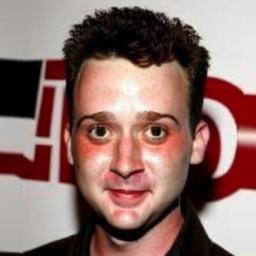} \\
  \end{tabular}
  \caption{Identity preservation: comparison across IP Adapter, PhotoMaker, InstantID, Arc2Face, and Diff-ID. All methods are driven by the same BLIP generated descriptive prompts for each subject.}
  \label{fig:identity_preservation_grid}
\end{figure}

Our comparative image grid in Figure~\ref{fig:id1_half_column} and the extended results in Figure~\ref{fig:identity_preservation_grid} show that Diff-ID preserves critical identity specific details, such as facial bone structure, eye spacing, nose shape, and lip contour, while also improving visual realism. IP Adapter and PhotoMaker exhibit weaker preservation in these examples, with noticeable drift in geometric structure and texture. Arc2Face maintains stronger identity cues but suffers from reduced sharpness and consistency across samples.

InstantID achieves the highest raw Face Similarity on both validation (75.13) and unseen data (74.12), as summarized in Table~\ref{tab:evaluation}. Diff-ID ranks closely behind in Face Similarity (72.68 and 71.53) while achieving substantially lower FID, indicating stronger realism under this evaluation. This trade off is summarized by the FIQ score in Equation~\eqref{eq:fiq}. Diff-ID attains the highest FIQ across both splits (71.74 on validation and 69.32 on unseen data), which supports the claim that Diff-ID offers the strongest FIQ based identity--realism trade off among the evaluated methods, not the highest raw identity similarity.
\subsubsection{Perceptual Realism}

Perceptual realism is measured by the Fréchet Inception Distance and is also reflected in the secondary FIQ score. Table~\ref{tab:evaluation} shows that Diff-ID achieves the lowest Fréchet Inception Distance on both the validation set (101.31) and unseen data (103.19), outperforming IP Adapter, PhotoMaker, InstantID, and Arc2Face on this realism metric. Because FIQ is defined as FS divided by FID, this realism advantage contributes directly to Diff-ID's highest FIQ scores, despite its lower Face Similarity compared to InstantID.

In contrast, PhotoMaker attains relatively low Fréchet Inception Distance (127.47 and 113.50) but has lower Face Similarity, which results in weaker FIQ. Arc2Face strikes a more balanced mid range performance, with solid Face Similarity but higher Fréchet Inception Distance than Diff-ID. These trends highlight that Diff-ID remains competitive in identity similarity while producing coherent skin textures, hair details, and backgrounds, avoiding the cartoon like or over smoothed artifacts present in some methods. We note that the Fréchet Inception Distance may overlook small yet perceptually salient flaws, while Face Similarity can over reward stylized similarity. Accordingly, the higher FIQ of Diff-ID should be interpreted as evidence of a favorable measured trade off between photorealism and identity retention across our benchmarks.

% ----------------------------------------------------------------------
% 5. Ablation Study
% ----------------------------------------------------------------------

\section{Ablation Study}

\subsection{Morphing}
\begin{figure*}[ht]
  \centering
  \setlength{\tabcolsep}{2pt}
  \renewcommand{\arraystretch}{1.0}
  \begin{tabular}{*{8}{>{\centering\arraybackslash}m{0.11\linewidth}}}

    \multicolumn{8}{c}{\bfseries DiffID Embedding Interpolation} \\[2pt]

    {\scriptsize \textbf{ID A}} & & & & & & & {\scriptsize \textbf{ID B}} \\[-2pt]

    \includegraphics[width=0.09\textwidth]{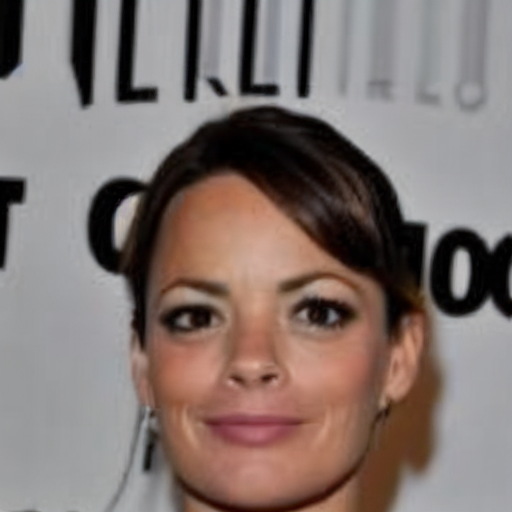} &
    \includegraphics[width=0.09\textwidth]{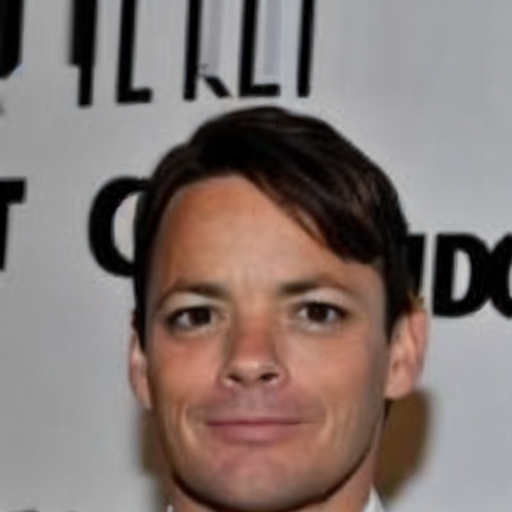} &
    \includegraphics[width=0.09\textwidth]{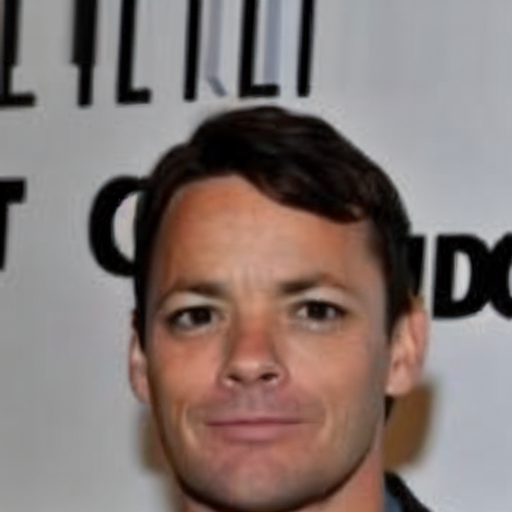} &
    \includegraphics[width=0.09\textwidth]{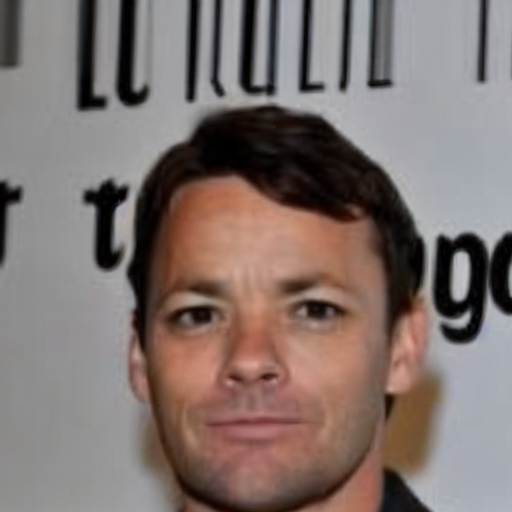} &
    \includegraphics[width=0.09\textwidth]{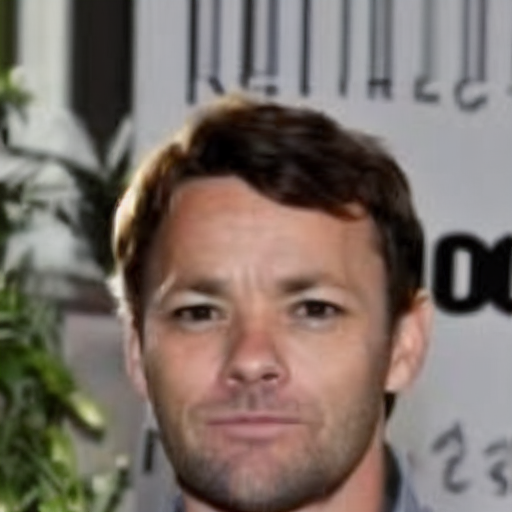} &
    \includegraphics[width=0.09\textwidth]{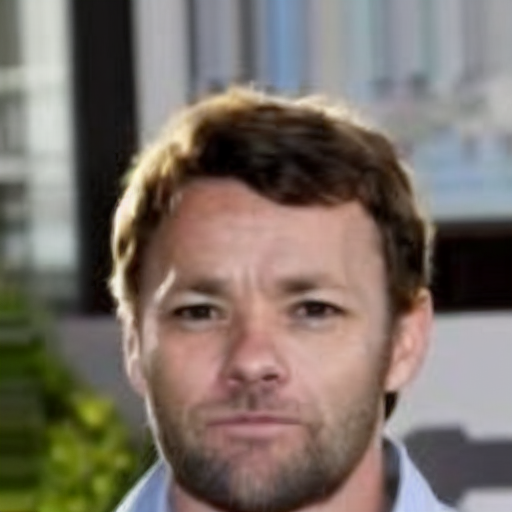} &
    \includegraphics[width=0.09\textwidth]{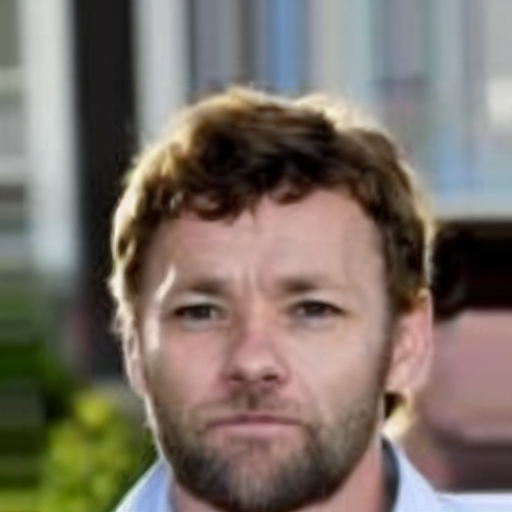} &
    \includegraphics[width=0.09\textwidth]{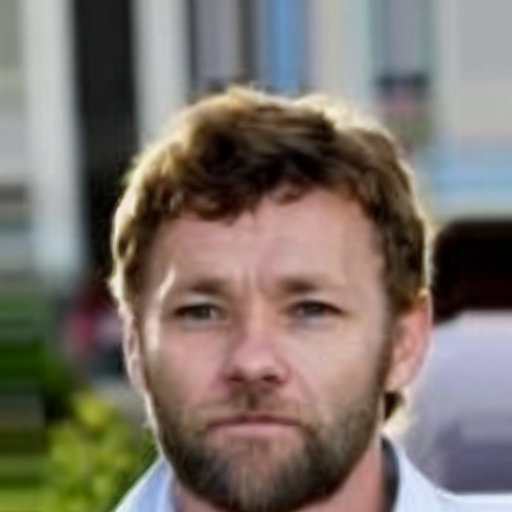} \\[-2pt]

    {\scriptsize 0.873 / 0.037} &
    {\scriptsize 0.683 / 0.159} &
    {\scriptsize 0.621 / 0.269} &
    {\scriptsize 0.524 / 0.409} &
    {\scriptsize 0.398 / 0.577} &
    {\scriptsize 0.259 / 0.725} &
    {\scriptsize 0.139 / 0.833} &
    {\scriptsize 0.055 / 0.876} \\[10pt]

    \multicolumn{8}{c}{\bfseries DiffID Morph (Linear Interpolation)} \\[4pt]
    \includegraphics[width=0.09\textwidth]{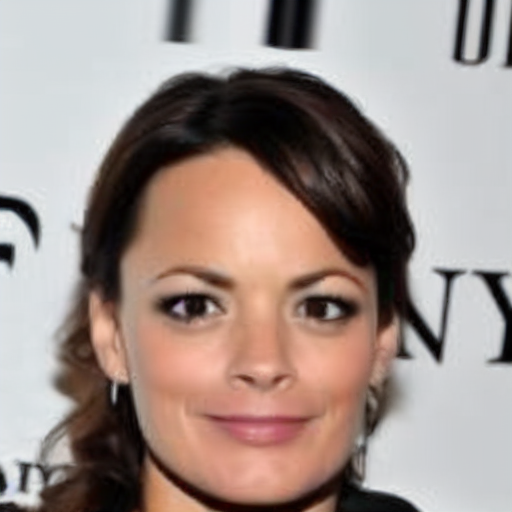} &
    \includegraphics[width=0.09\textwidth]{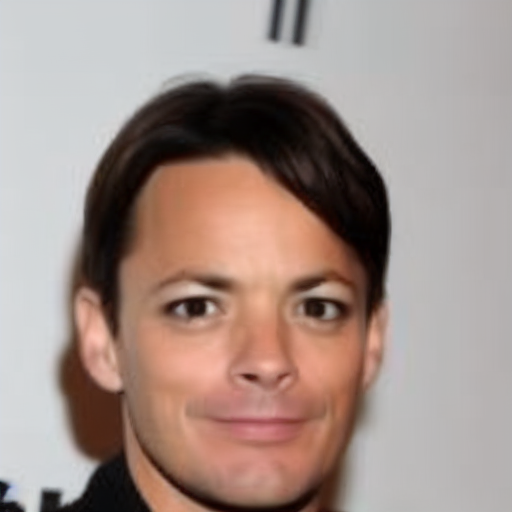} &
    \includegraphics[width=0.09\textwidth]{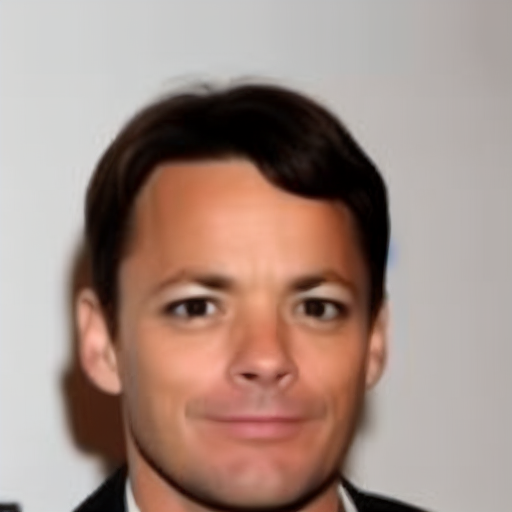} &
    \includegraphics[width=0.09\textwidth]{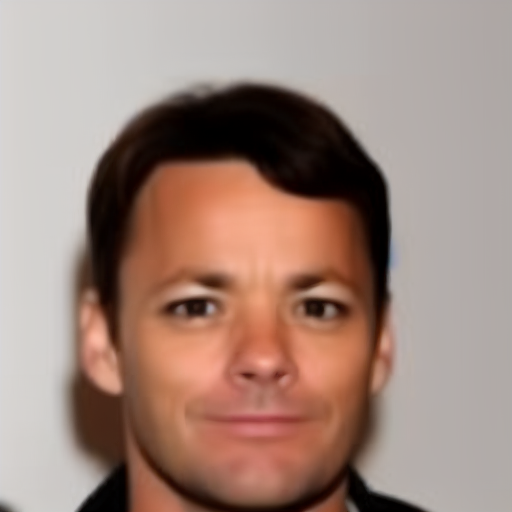} &
    \includegraphics[width=0.09\textwidth]{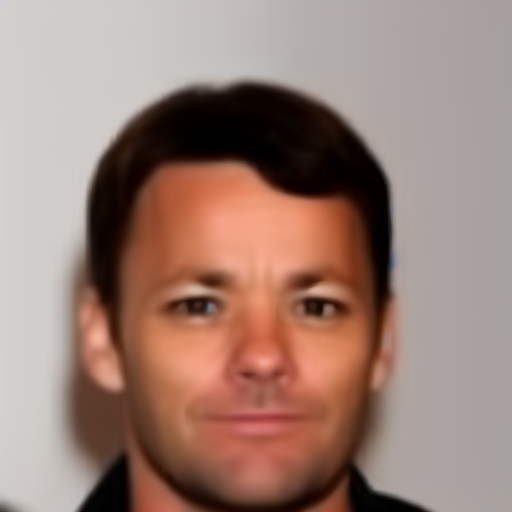} &
    \includegraphics[width=0.09\textwidth]{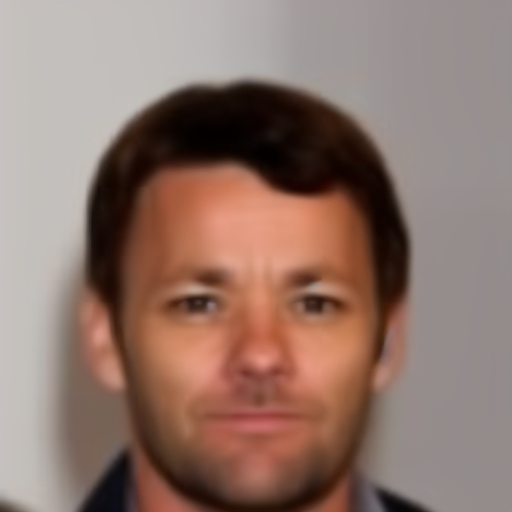} &
    \includegraphics[width=0.09\textwidth]{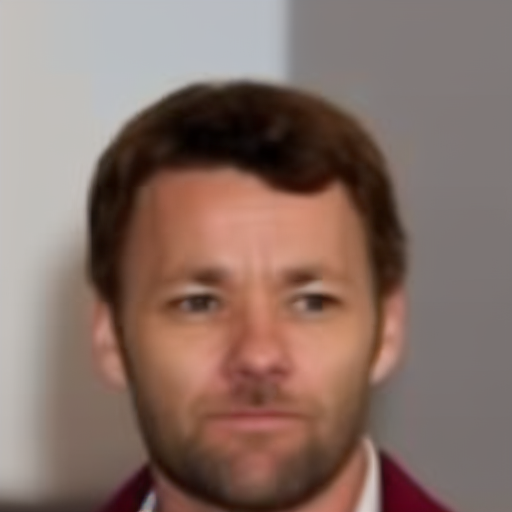} &
    \includegraphics[width=0.09\textwidth]{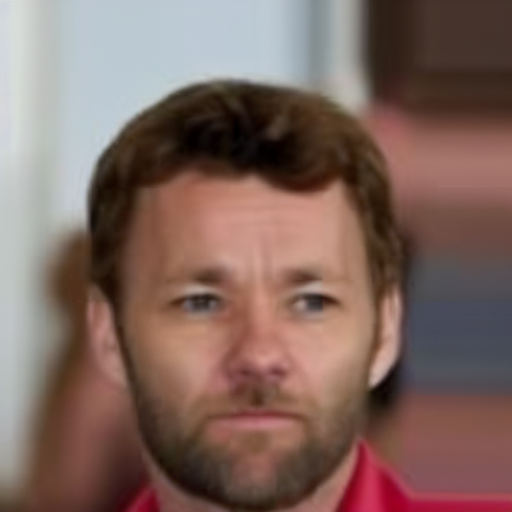} \\[-2pt]

    {\scriptsize 1.000 / 0.021} &
    {\scriptsize 0.772 / 0.154} &
    {\scriptsize 0.689 / 0.269} &
    {\scriptsize 0.580 / 0.419} &
    {\scriptsize 0.427 / 0.575} &
    {\scriptsize 0.273 / 0.693} &
    {\scriptsize 0.147 / 0.813} &
    {\scriptsize 0.062 / 1.000} \\[10pt]

    \multicolumn{8}{c}{\bfseries DiffID Morph (Spherical Interpolation)} \\[4pt]
    \includegraphics[width=0.09\textwidth]{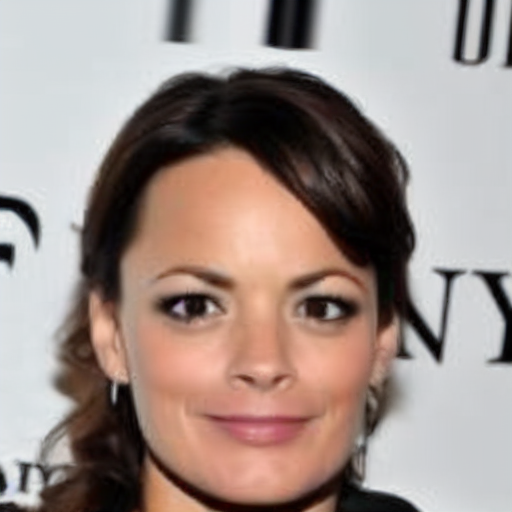} &
    \includegraphics[width=0.09\textwidth]{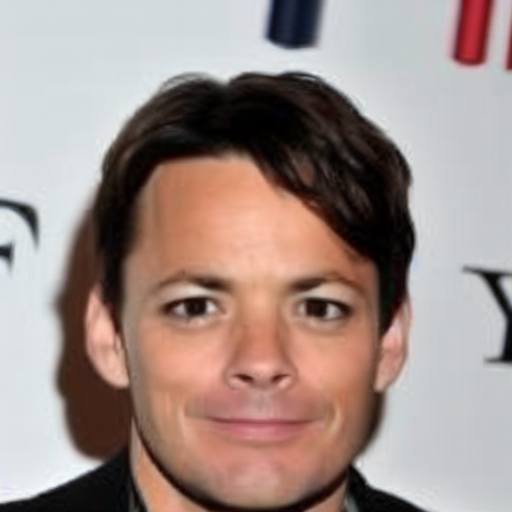} &
    \includegraphics[width=0.09\textwidth]{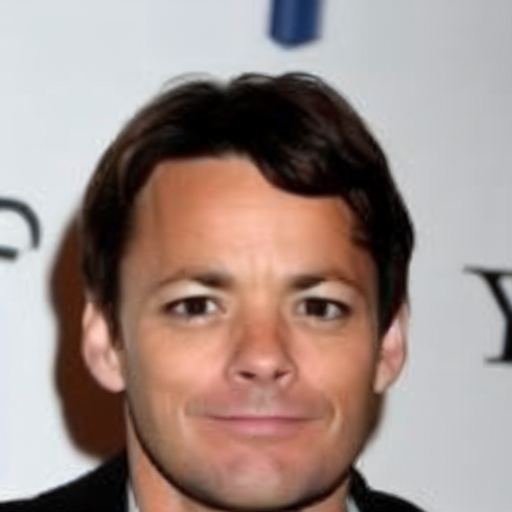} &
    \includegraphics[width=0.09\textwidth]{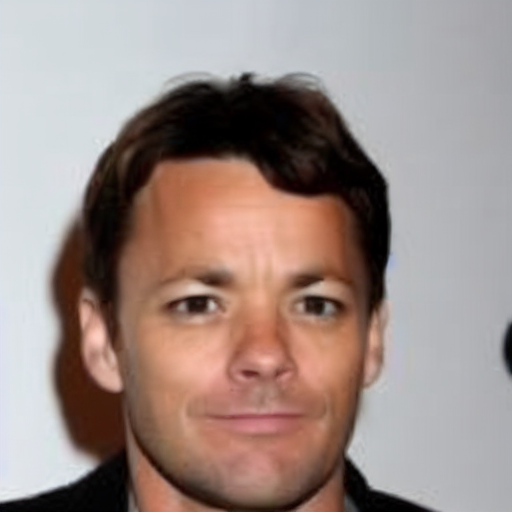} &
    \includegraphics[width=0.09\textwidth]{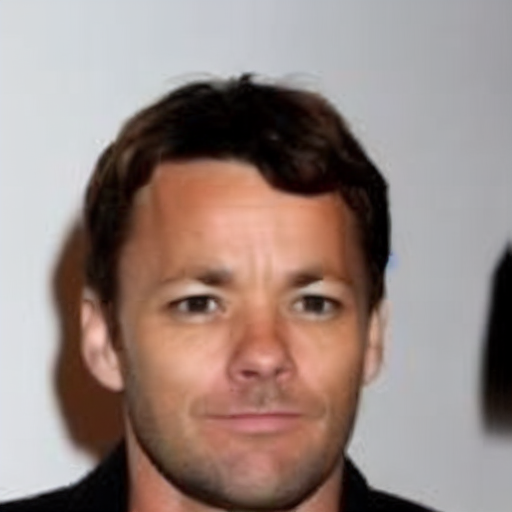} &
    \includegraphics[width=0.09\textwidth]{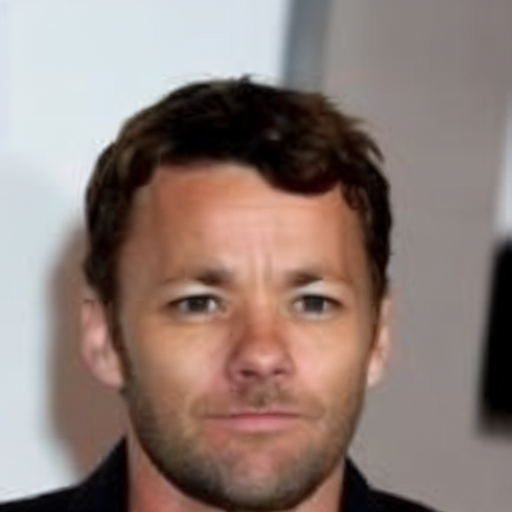} &
    \includegraphics[width=0.09\textwidth]{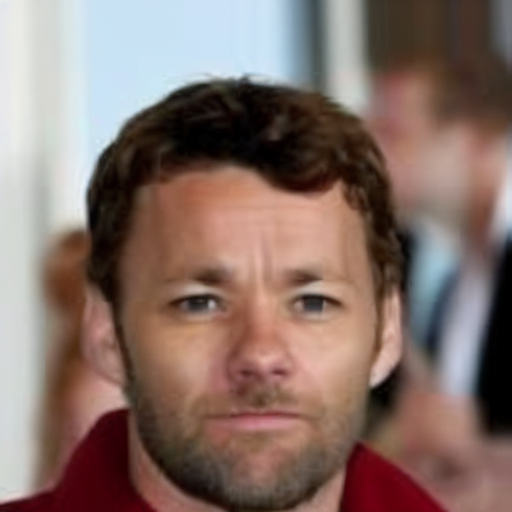} &
    \includegraphics[width=0.09\textwidth]{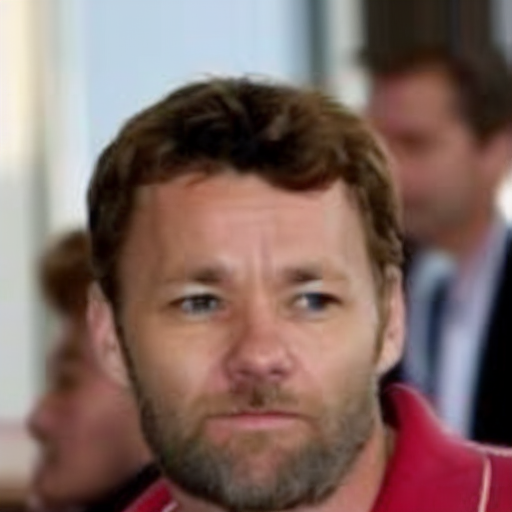} \\[-2pt]

    {\scriptsize 1.000 / 0.021} &
    {\scriptsize 0.734 / 0.182} &
    {\scriptsize 0.639 / 0.320} &
    {\scriptsize 0.551 / 0.428} &
    {\scriptsize 0.418 / 0.566} &
    {\scriptsize 0.306 / 0.681} &
    {\scriptsize 0.201 / 0.786} &
    {\scriptsize 0.018 / 1.000} \\

  \end{tabular}
  \caption{Comparison of morphing methods. Top: embedding space interpolation. Middle: linear interpolation. Bottom: spherical interpolation. Values under each image are shown as $(s_A / s_B)$, where $s_A$ is ArcFace cosine similarity to identity A (left endpoint) and $s_B$ is similarity to identity B (right endpoint).}
  \label{fig:morph-grid}
\end{figure*}

Figure~\ref{fig:morph-grid} illustrates three morphing strategies: (i) embedding only interpolation of fused identity vectors, (ii) DiffID Morph via linear interpolation, and (iii) DiffID Morph via spherical interpolation.

Embedding only interpolation (top row) produces overly smooth, low detail transitions that often misalign key facial features and lack high frequency texture. By contrast, DiffID linear interpolation (middle row) injects identity and semantic cues at each denoising step, yielding sharp, coherent blends, though some intermediate frames still exhibit mild ghosting of features. DiffID spherical interpolation (bottom row) further regularizes interpolation on the hypersphere, which improves mid point consistency in bone structure, skin texture, and overall likeness.

Although we do not report quantitative FIQ scores for each frame here, the qualitative visual fidelity of DiffID with spherical interpolation is consistent with the identity--realism trade off observed in our main evaluation. We therefore present these morphing results as qualitative evidence of smooth interpolation behavior rather than as a complete biometric or aggregate morphing evaluation. We acknowledge that pure Face Similarity on individual frames can be inflated by stylized or avatar like artifacts; these visual results suggest that our diffusion based joint interpolation approach can produce smooth, photorealistic face morphs without the need for per identity fine tuning or multiple checkpoints.

\subsection{Impact of Different Adapter Strategies}
\begin{figure}[ht]
    \centering
    \setlength{\tabcolsep}{1pt}
    \begin{tabular}{ccccc}
        \textbf{Original} &
        \includegraphics[width=0.14\columnwidth]{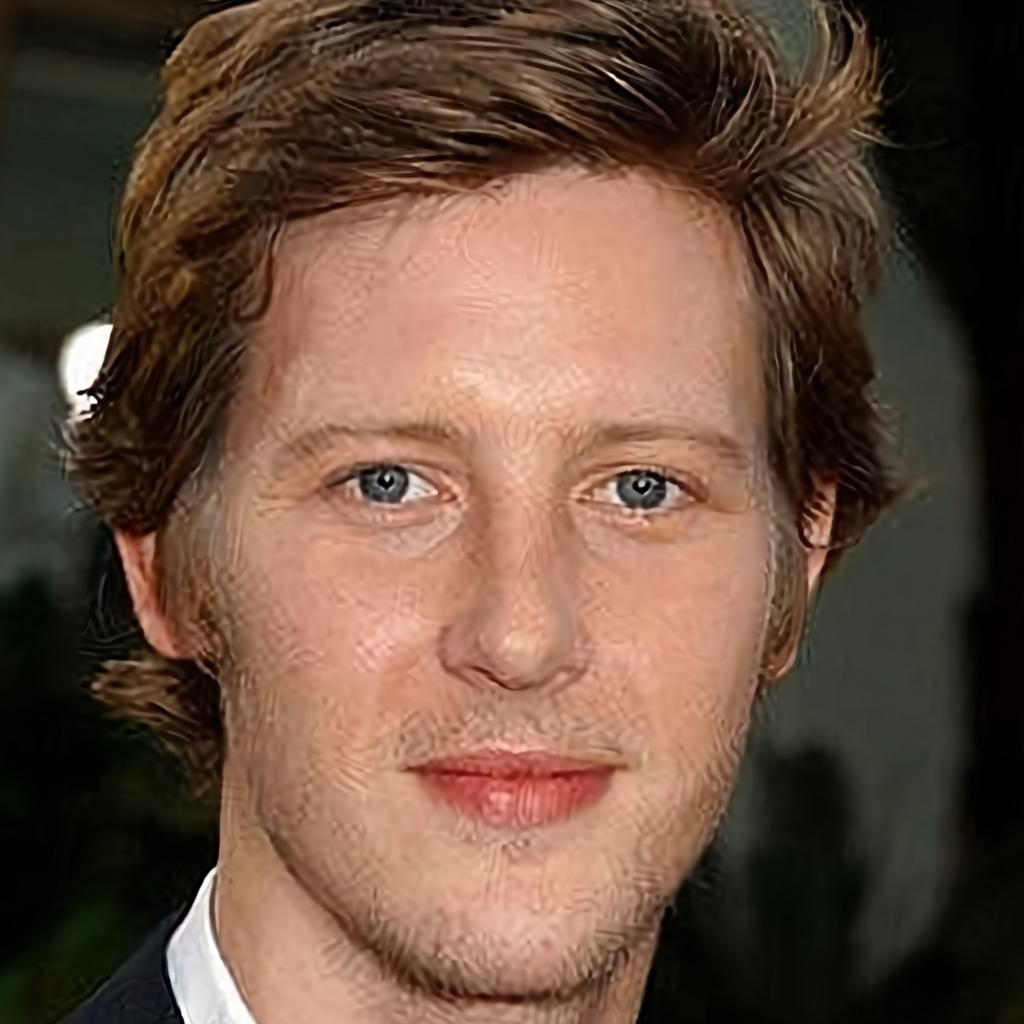} &
        \includegraphics[width=0.14\columnwidth]{Ablation/orignalm.png} &
        \includegraphics[width=0.14\columnwidth]{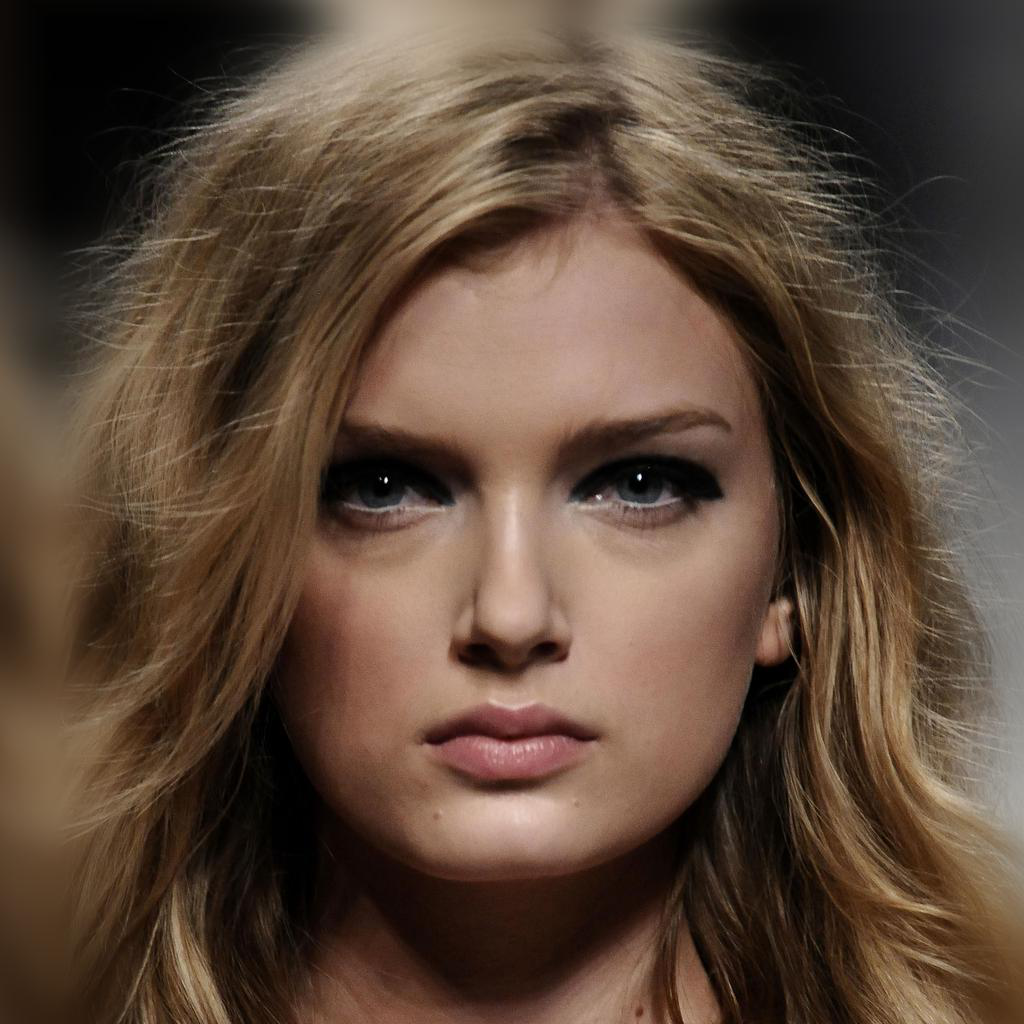} &
        \includegraphics[width=0.14\columnwidth]{Ablation/orignalw.png} \\[1pt]

        \textbf{DiffID ArcFace} &
        \includegraphics[width=0.14\columnwidth]{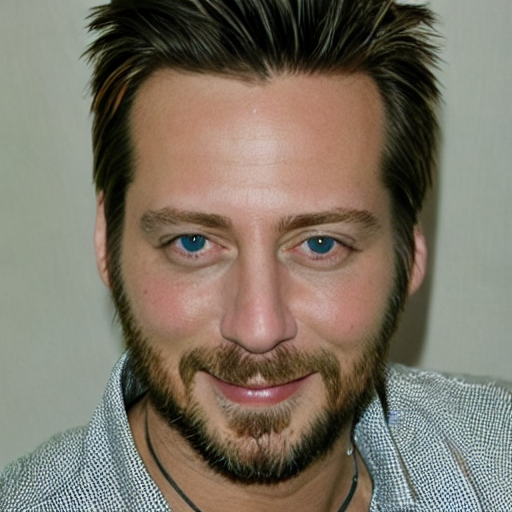} &
        \includegraphics[width=0.14\columnwidth]{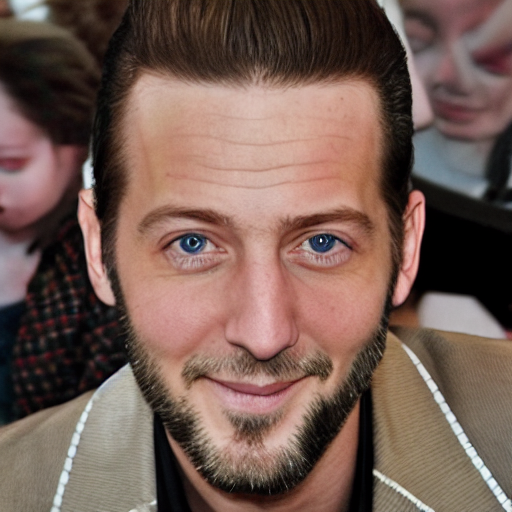} &
        \includegraphics[width=0.14\columnwidth]{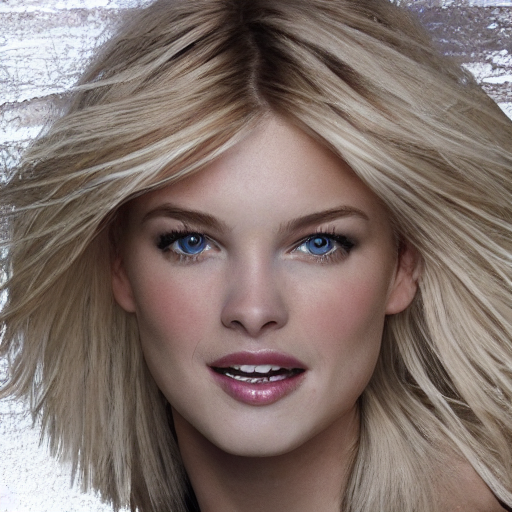} &
        \includegraphics[width=0.14\columnwidth]{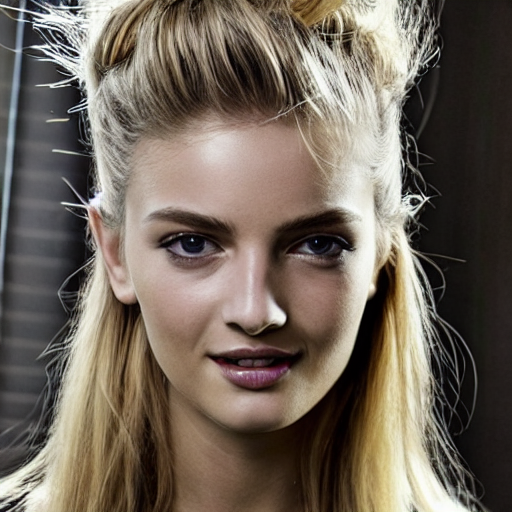} \\[1pt]

        \textbf{DiffID CLIP} &
        \includegraphics[width=0.14\columnwidth]{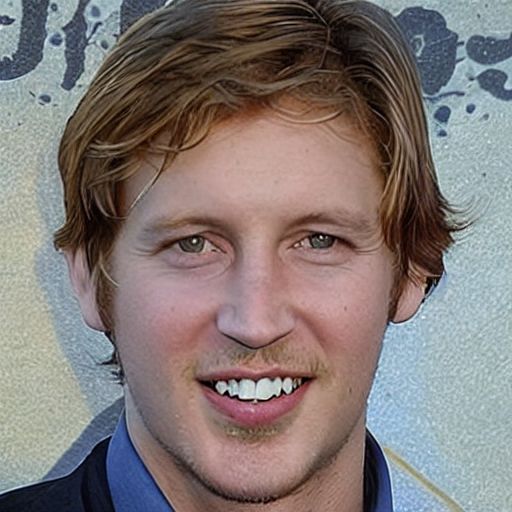} &
        \includegraphics[width=0.14\columnwidth]{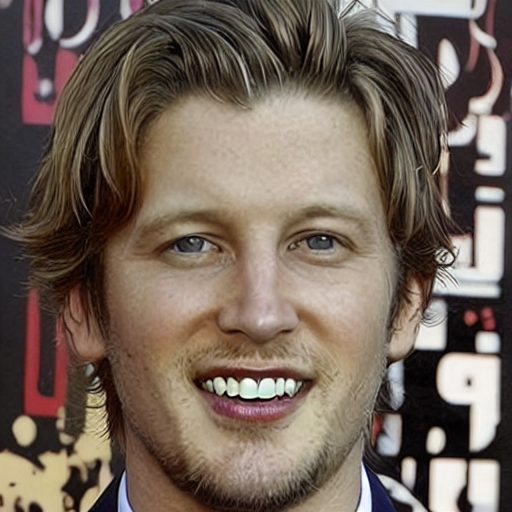} &
        \includegraphics[width=0.14\columnwidth]{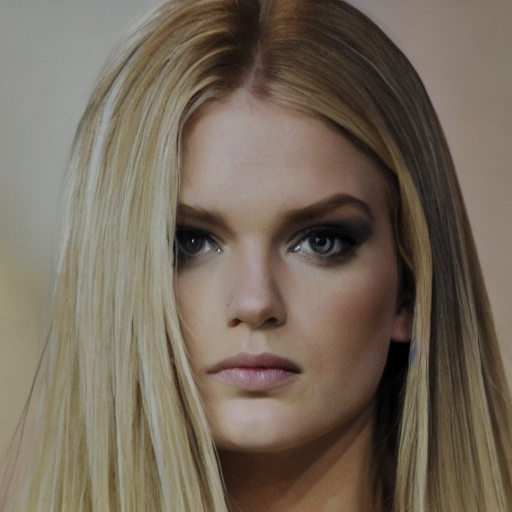} &
        \includegraphics[width=0.14\columnwidth]{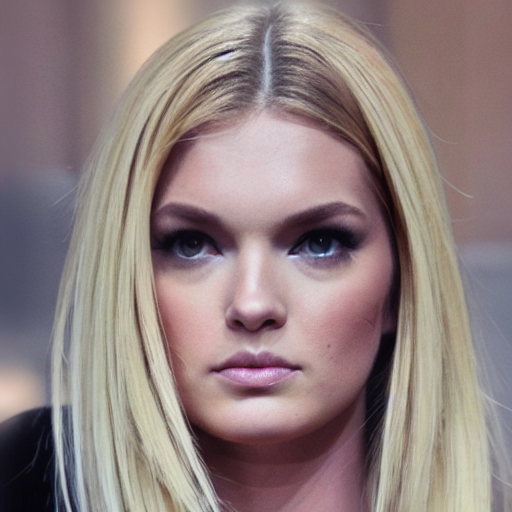} \\[1pt]

        \textbf{DiffID Joint} &
        \includegraphics[width=0.14\columnwidth]{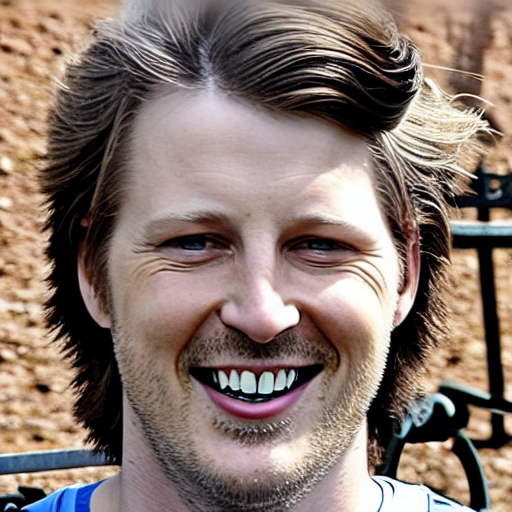} &
        \includegraphics[width=0.14\columnwidth]{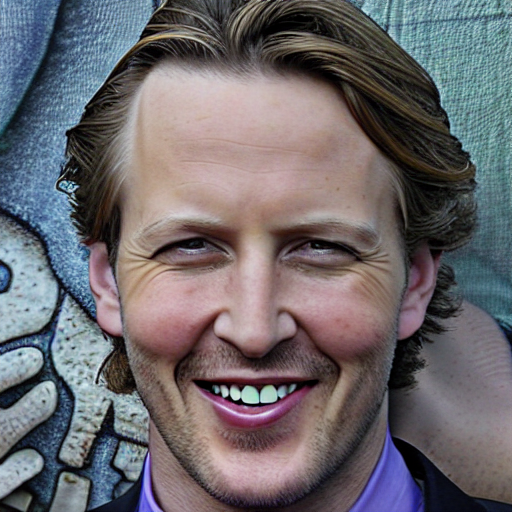} &
        \includegraphics[width=0.14\columnwidth]{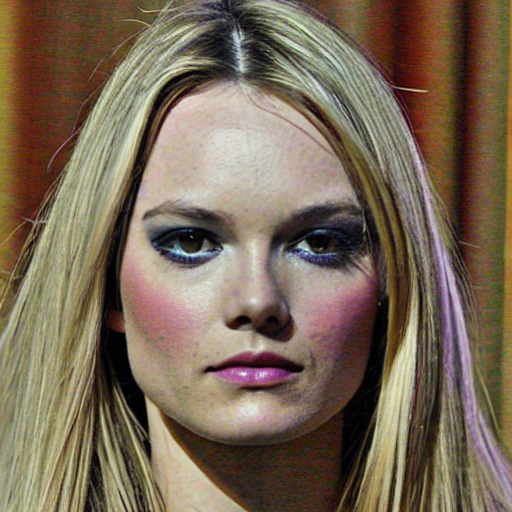} &
        \includegraphics[width=0.14\columnwidth]{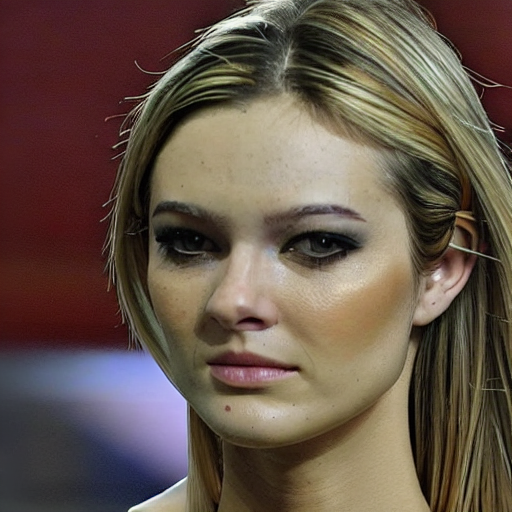} \\[1pt]

        \textbf{DiffID Final} &
        \includegraphics[width=0.14\columnwidth]{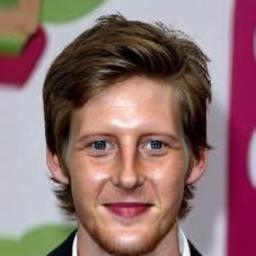} &
        \includegraphics[width=0.14\columnwidth]{Ablation/Finalm2.jpeg} &
        \includegraphics[width=0.14\columnwidth]{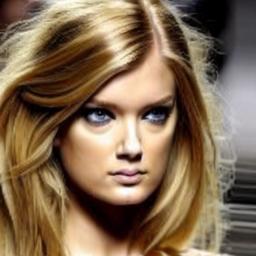} &
        \includegraphics[width=0.14\columnwidth]{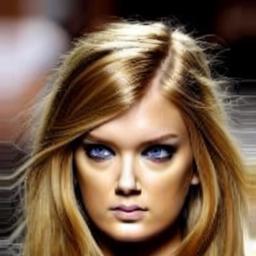} \\
    \end{tabular}
    \caption{Adapter ablation: identity preservation across variants. DiffID Joint combines ArcFace and CLIP image embeddings through dual cross attention but omits the Fusion multilayer perceptron, while DiffID Final adds the Fusion multilayer perceptron.}
    \label{fig:adapter-ablation}
\end{figure}

Figure~\ref{fig:adapter-ablation} visualizes outputs from four adapter variants: DiffID ArcFace, which uses only ArcFace embeddings without dual cross attention; DiffID CLIP, which uses only CLIP embeddings without dual cross attention; DiffID Joint, which combines ArcFace and CLIP embeddings using dual cross attention but excludes the Fusion multilayer perceptron; and DiffID Final, which adds the Fusion multilayer perceptron on top of the Joint variant. Each row shows two representative identities.

\paragraph{Qualitative Observations}
The ArcFace only adapter (DiffID ArcFace) preserves broad geometry (jawline, eye spacing) but lacks fine texture detail. The CLIP only adapter (DiffID CLIP) recovers richer semantic context and sharper textures but introduces identity drift, for example nose shape and lip contour deviate from the source. The Joint adapter (DiffID Joint) uses both ArcFace and CLIP embeddings with dual cross attention, which balances geometry and texture but still shows slight mid face averaging. Because this variant omits the Fusion multilayer perceptron, the comparison between DiffID Joint and DiffID Final isolates the additional effect of the nonlinear Fusion MLP refinement. Our final DiffID Final adapter visually improves high frequency details (for example pores and subtle wrinkles) while maintaining bone structure, lip shape, and eye geometry across examples.
\begin{table}[ht]
  \centering
  \begingroup
  \small
  \setlength{\tabcolsep}{3pt}
  \renewcommand{\arraystretch}{1.0}
    \begin{tabular}{l c}
    \hline
    \textbf{Model Variant} & \textbf{FS $\uparrow$} \\
    \hline
    DiffID ArcFace & 48.91 \\
    DiffID CLIP    & 53.41 \\
    DiffID Joint   & 65.83 \\
    DiffID Final   & \textbf{72.68} \\
    \hline
  \end{tabular}
  \caption{Adapter ablation results (Face Similarity). The final configuration combining ArcFace and CLIP with dual cross attention and a Fusion multilayer perceptron corresponds to the Diff-ID result in Table~\ref{tab:evaluation}.}
  \label{tab:abface_similarity}
  \endgroup
\end{table}
\paragraph{Quantitative Insights}

As shown in Table~\ref{tab:abface_similarity}, the ArcFace only adapter scores 48.91, which reflects limited textural fidelity, while the CLIP only variant achieves 53.41 by leveraging semantic detail but misaligning fine identity cues. Both of these single embedding variants omit dual cross attention. The Joint adapter reaches 65.83 by combining ArcFace and CLIP embeddings through dual cross attention, indicating that cross modal identity--semantic interaction improves Face Similarity over either embedding alone. Finally, DiffID Final reaches 72.68 after adding the Fusion multilayer perceptron, suggesting that the MLP further refines the jointly attended representation. These ablations support the usefulness of both ArcFace and CLIP embeddings, the dual cross attention fusion stage, and the final Fusion MLP, although Table~\ref{tab:evaluation} shows that InstantID remains higher in raw Face Similarity among the external baselines.

\subsection{Impact of Identity Loss}

\subsubsection*{Convergence Behavior}
\begin{figure}[ht]
  \centering
  \includegraphics[width=0.6\columnwidth]{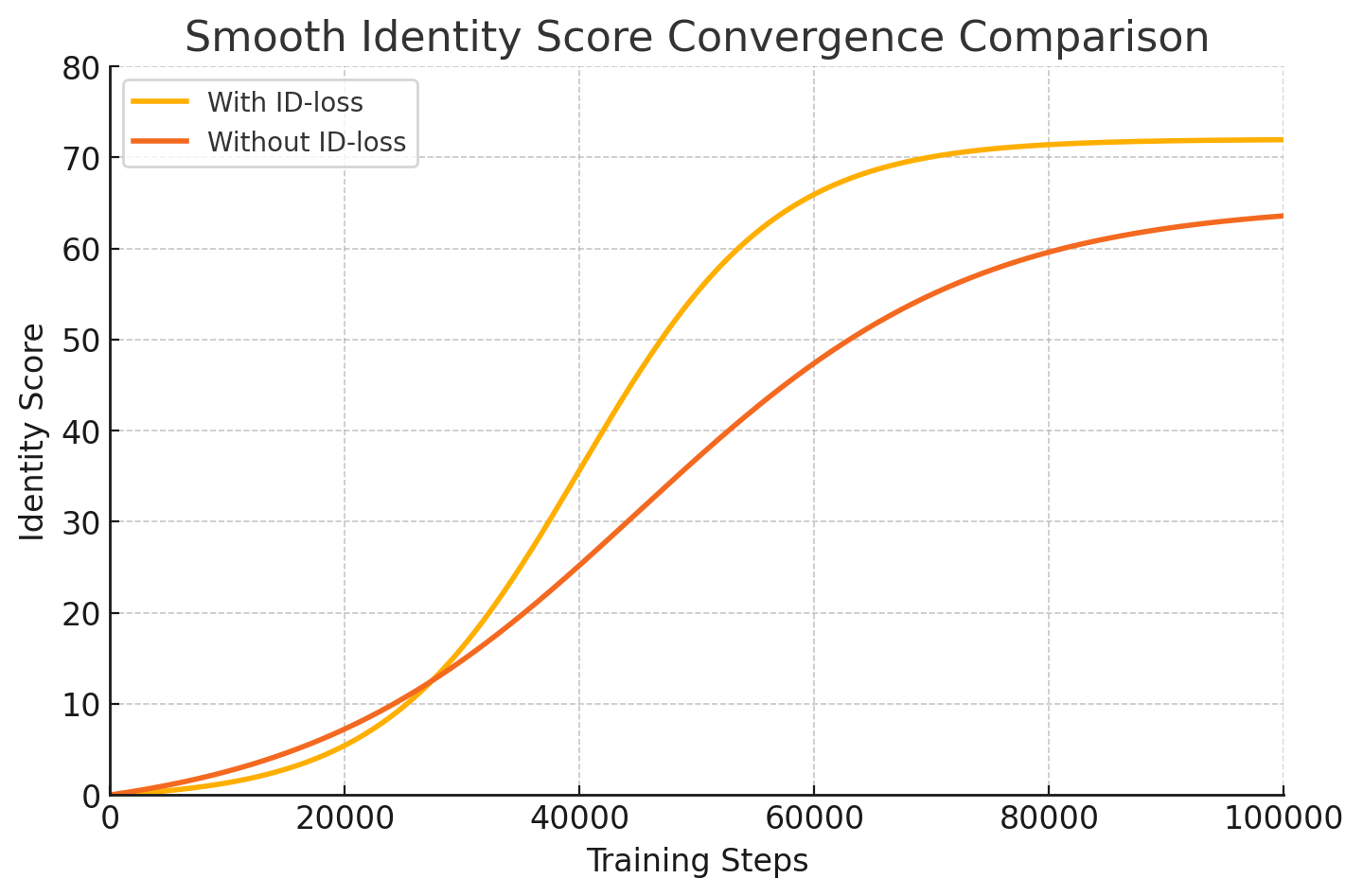}
  \caption{Training dynamics with and without the identity loss term. Exponential timestep weighting concentrates identity supervision on later denoising stages, which stabilizes convergence and improves identity fidelity.}
  \label{fig:loss_curve}
\end{figure}

Figure~\ref{fig:loss_curve} compares training curves for the denoising loss when using our pseudo discriminator identity loss versus a baseline without it. Incorporating the identity term yields faster early convergence in this training curve: the model learns to preserve coarse identity features earlier and minimizes the combined loss, whereas the baseline drifts more gradually as it struggles to infer identity from reconstruction alone. By emphasizing identity coherence at low noise levels, our weighted identity loss is intended to guide the network toward stable identity retention from the outset. This experiment directly ablates the identity loss: the model trained with the ArcFace identity loss reaches a higher final identity score and converges faster than the model trained without it. Since the identity loss is designed to affect identity preservation rather than image distribution realism, we report this ablation using identity score and convergence behaviour rather than FID or FIQ.

\subsubsection*{Effect of Identity Loss Weighting}
\begin{figure}[ht]
  \centering
  \setlength{\tabcolsep}{2pt}
  \begin{tabular}{cccccc}
    \textbf{Original} &
    \(\lambda=0.10\) &
    \(\lambda=0.20\) &
    \(\lambda=0.30\) &
    \(\lambda=0.40\) &
    \(\lambda=0.50\) \\[3pt]

    \includegraphics[width=0.122\columnwidth,height=2.6cm,keepaspectratio]{ID1.png} &
    \includegraphics[width=0.13\columnwidth,height=2.6cm,keepaspectratio]{DiffidID1.jpeg} &
    \includegraphics[width=0.13\columnwidth,height=2.6cm,keepaspectratio]{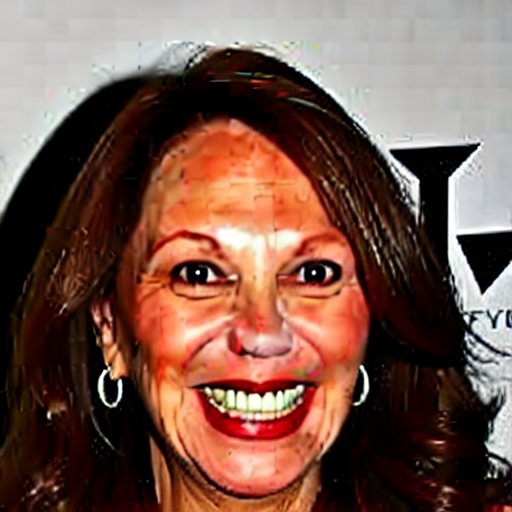} &
    \includegraphics[width=0.13\columnwidth,height=2.6cm,keepaspectratio]{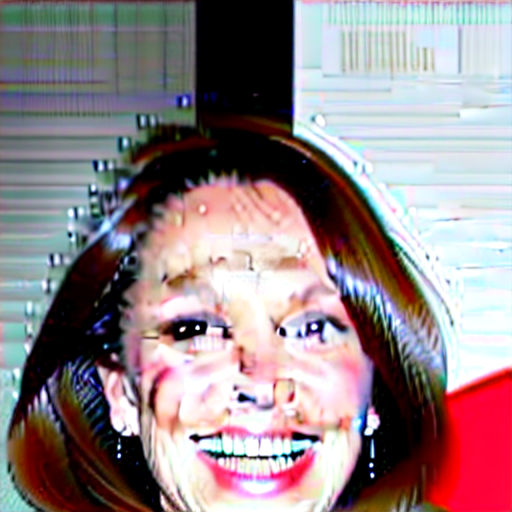} &
    \includegraphics[width=0.13\columnwidth,height=2.6cm,keepaspectratio]{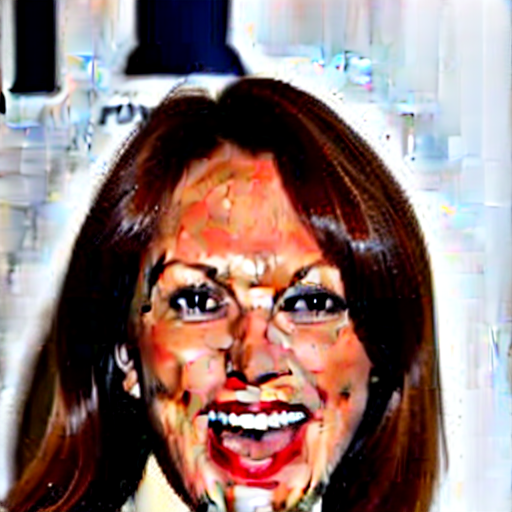} &
    \includegraphics[width=0.13\columnwidth,height=2.6cm,keepaspectratio]{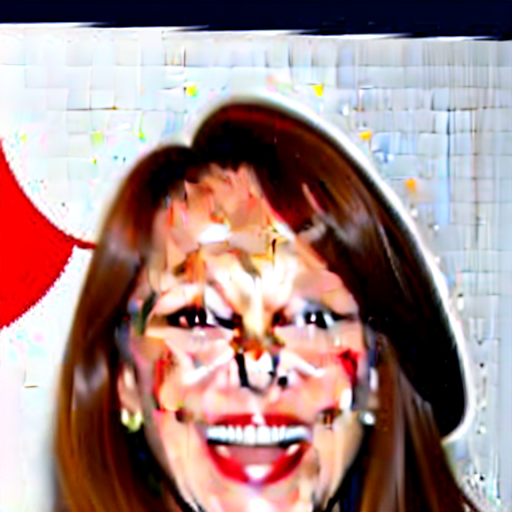} \\
  \end{tabular}
  \caption{Effect of identity regularization strength (\(\lambda\)) on generated outputs. Moderate values achieve a good balance between identity fidelity and perceptual quality.}
  \label{fig:lambda_identity_loss}
\end{figure}

Figure~\ref{fig:lambda_identity_loss} illustrates the impact of varying the identity loss weight \(\lambda\). At low values (\(\lambda=0.10\)), the identity term gently steers the diffusion process, which yields strong identity retention without visibly disturbing the denoising dynamics. As \(\lambda\) increases to values between 0.30 and 0.50, the model over prioritizes identity, which causes artifacts and degraded denoising quality. Features become unnaturally sharp or appear stuck, and background details collapse. These results suggest that a moderate weighting (around 0.10 to 0.20) achieves the best balance between identity fidelity and photorealism within this ablation, and they motivate our choice of \(\lambda=0.10\) in Equation~\eqref{eq:loss_total}. This sweep is intended as a qualitative ablation of identity-loss strength. Low to moderate identity-loss weights preserve identity without introducing visible artifacts, whereas larger weights over prioritize identity and damage realism. We therefore use \(\lambda=0.10\) in the final model. We do not report FID or FIQ for this sweep because the purpose of the figure is to show the visual failure mode caused by excessive identity regularization.

% ----------------------------------------------------------------------
% 6. Limitations and Future Work
% ----------------------------------------------------------------------

\section{Limitations and Future Work}
While Diff-ID shows a favorable identity--realism trade off for identity consistent face synthesis and qualitative morphing, several limitations remain. First, by relying on frozen CLIP image and text encoders for semantic context, our framework lacks explicit control over non facial elements, such as backgrounds, clothing, and accessories, which remain governed by coarse CLIP embeddings rather than dedicated spatial or attribute specific modules. Second, Diff-ID is optimized for single, frontal face crops; extreme head poses, severe occlusions, or full body scenes can degrade identity fidelity, since the model has not been trained to disentangle complex scene elements or non facial regions. Third, our morphing pipeline employs simple linear and spherical interpolation of latents and identity embeddings, which, while effective for smooth transitions, does not support attribute conditioned or region specific morph trajectories (for example selectively blending expressions or hairstyles).  

Although we instantiate Diff-ID on a UNet backbone for comparability with existing identity guided diffusion models, the dual cross attention adapter operates purely on projected token sequences and cross attention blocks. As such, it is architecturally compatible with transformer based diffusion backbones such as DiT. Adapting the method would primarily involve choosing appropriate injection points within DiT attention layers rather than redesigning the adapter itself. Exploring Diff-ID style identity conditioning in DiT architectures is therefore an interesting direction for future work.

Finally, our morphing experiments are designed to probe the identity stability of the generator rather than to evaluate biometric vulnerability. We therefore do not report aggregate morphing metrics, attack specific metrics such as false match rates, false non match rates, or morph attack detection scores, or comparisons with dedicated morphing baselines. These evaluations depend on external verification systems and standardized protocols. Studying Diff-ID morphs under biometric security and morph attack detection benchmarks is orthogonal to our current focus and remains an important avenue for future research.

The same capabilities also raise broader impact concerns. Identity consistent face generation and morphing can support synthetic data creation, privacy preserving evaluation, and biometric robustness testing, but they can also be misused for impersonation, deepfakes, identity fraud, or morph attacks against biometric systems. Any release of trained models or generated identity data should therefore consider dataset consent and licensing, demographic bias analysis, usage restrictions, provenance or watermarking mechanisms, and limits on arbitrary identity cloning.

Future research can also address the limitations above by integrating ControlNet style spatial conditioning or learned attention masks to separately modulate background and clothing attributes, thereby enriching non facial control. Enhancing the dataset with multi view and occluded face samples, or incorporating three dimensional aware diffusion priors, could improve robustness to pose and occlusion. On the morphing side, developing learned interpolation networks or diffusion paths guided by semantic anchors would enable finer grained, attribute aware transitions, for example expression morphing without altering identity geometry. By extending Diff-ID in these directions, one can build a more generalizable and controllable platform for secure, identity anchored image synthesis and morphing.

% ----------------------------------------------------------------------
% 7. Conclusion
% ----------------------------------------------------------------------

\section{Conclusion}
In this work, we introduced \textbf{Diff-ID}, a unified diffusion framework for high resolution facial image generation and morphing that explicitly encourages identity consistency. By fusing ArcFace and CLIP embeddings through a lightweight dual cross attention adapter within a fine tuned Stable Diffusion UNet, and by incorporating a pseudo discriminator identity loss with exponential timestep weighting, Diff-ID achieves competitive Face Similarity together with the lowest FID among the evaluated methods. Our evaluations on held out and unseen face sets show that InstantID obtains the highest raw Face Similarity, while Diff-ID obtains the best FID and FIQ, supporting a stronger identity--realism trade off rather than superior raw identity retention. Furthermore, our DDIM based morphing pipeline provides qualitative smooth, photorealistic face interpolations without per subject fine tuning or multiple checkpoints.  
 
Diff-ID lays the groundwork for further study of identity anchored facial synthesis, privacy preserving data augmentation, and controlled morphing. Addressing the identified limitations, including non facial attribute control, transformer based diffusion backbones, reproducible evaluation protocols, continued reporting of FS and FID as primary metrics, and more sophisticated morphing benchmarks, will be essential for establishing the robustness and responsible use of identity anchored diffusion models.

\bibliographystyle{unsrt}
\bibliography{main}

\end{document}

%% file: math_commands.tex
\usepackage{amsmath,amsfonts,bm}

\def\eqref#1{equation~\ref{#1}}
\def\1{\bm{1}}

\DeclareMathAlphabet{\mathsfit}{\encodingdefault}{\sfdefault}{m}{sl}
\SetMathAlphabet{\mathsfit}{bold}{\encodingdefault}{\sfdefault}{bx}{n}